\documentclass[letterpaper, 10 pt, conference]{ieeeconf}

\IEEEoverridecommandlockouts                              

\usepackage{times,enumerate} 
 \usepackage[usenames,dvipsnames,svgnames,table]{xcolor}

\usepackage{subfloat}
\usepackage{hyperref}

\usepackage{graphicx}
\usepackage{setspace}
\usepackage{bbm}
\usepackage{mathdots,mathrsfs}
\usepackage{amssymb,latexsym,amsfonts,amsmath,cite,comment,relsize}
\usepackage{stmaryrd}
\usepackage{caption}
\usepackage[dvips]{psfrag}
\usepackage{setspace}
\usepackage{amsmath}
\usepackage{url}
\usepackage{soul}
\usepackage{blindtext}
\usepackage{graphicx}
\usepackage{lipsum}

\usepackage{adjustbox}
\usepackage{tikz}
\usepackage{pgfplots}
\pgfplotsset{every tick label/.append style={font=\small}}

\usepackage{resizegather}

\usepackage{relsize}
\usetikzlibrary{patterns}

\usepackage{epsfig} 
\usepackage{subfig}
\usepackage{algorithm,algpseudocode}

\newtheorem{remark}{Remark}

\tikzstyle{vertex}=[circle, draw, inner sep=0pt, minimum size=6pt]

\usepackage[sans]{dsfont}
\usepackage[font={small,it}]{caption}

\newcommand{\suchthat}{\;\ifnum\currentgrouptype=16 \middle\fi|\;}

\newcommand{\R}{\mathbb{R}}

\usepackage{algorithm,algpseudocode}

\title{\LARGE \bf
	 A Layered Architecture for Active Perception: \\ Image Classification using Deep Reinforcement Learning     
}

\author{Hossein K. Mousavi, Guangyi Liu, Weihang Yuan, Martin Tak\'a\u{c},  H\'ector Mu\~noz-Avila,  and Nader Motee$^{1}$
    \thanks{$^{1}$H.K.M., G.L., and N.M. are with the Department of Mechanical Engineering and Mechanics, Lehigh University,
        Bethlehem, PA 18015, USA
        {\tt\small \{mousavi,gul316, motee\}@lehigh.edu}. W.Y. and H.M. are with the Department of Computer Science and Engineering, Lehigh University {\tt\small \{hem4, wey218\}@lehigh.edu}. M.T. is with the Department of Industrial and Systems Engineering, Lehigh University {\tt\small \{takac.mt\}@gmail.com}. 
}%
}

\usepackage{xargs}                      

\begin{document}
	
	\maketitle
	\thispagestyle{empty}
	\pagestyle{empty}

\begin{abstract}  
	
	We propose a planning and perception mechanism for a robot (agent), that can only observe the underlying environment partially, in order to solve an image classification problem. A three-layer architecture  is suggested that consists of a meta-layer that decides the intermediate goals, an {action}-layer that selects local actions as the agent navigates {towards} a goal, and a classification-layer that evaluates the reward and makes a prediction. We design and implement these layers using deep reinforcement learning. A generalized policy gradient algorithm is utilized to learn the parameters of these layers to maximize the expected reward. Our proposed methodology is tested on {the} MNIST dataset of handwritten digits, {which provides us with a level of explainability while interpreting the  agent's intermediate goals and course of action}. 
\end{abstract}


	\section{Introduction}
	
There has been a rapidly growing  interest in goal reasoning in recent years; planning mechanisms for  {agents} that {are} capable of explicitly reasoning about their goals and changing them whenever it becomes  necessary  \cite{aha2018goal,munoz2018adaptive}. The  potential applications of goal reasoning spans over several  research fields, for example, only to name a few, controlling underwater unmanned vehicles \cite{wilson2018goal}, playing digital games \cite{dannenhauer2015goal}, and air combat simulations \cite{floyd2017goal}.

	 {One of the promising recent frameworks for goal-based planning and reasoning} is hierarchical {deep Q-networks} (hDQN) \cite{kulkarni2016hierarchical}, {which} consists of two layers: a meta-layer that plans strategically and an action-layer that plans local navigation. The meta-layer receives a state as its input and outputs a goal, a condition that can be evaluated in a given state. The action-layer receives a state and a goal as its input. Then, it selects and executes actions until the agent reaches a state where the goal is achieved. Both layers use a deep neural network similar to that of DQN with some important differences: the meta-layer selects goals in order to maximize external rewards from the environment, while the action-layer selects actions to maximize designer-defined intrinsic rewards (e.g., $1$ for reaching the goal state and $0$ otherwise).
	
{In this work, we consider the problem of  exploring an environment by a robot for classification purposes. Contrary to the standard assumptions made in the literature, we assume that robot can only partially observe the environment, where each observation depends on  the actions taken by the robot.  The first and second layers of our proposed architecture  {are similar to those} of hDQN, while the third layer perform a classification task and evaluates the reward in a differentiable manner.} 

{Our approach has other differences from hDQN. First,}	note that in hDQN requirement, the action-layer reaches a state achieving the goal. However, find that this assumption  is too restrictive, {unnecessary, and potentially unrealizable due to partial observability} for our purposes. Instead, our method relaxes this requirement by allowing a robot to move a few steps towards the goal, but not necessarily reaching to it. This flexibility is needed because our intrinsic objective is to explore the  environment. Therefore, the goal planner {should only dictate} a desired general direction of exploration rather than imposing a hard constraint to reach a specific {position}. In this sense, our goals play a similar role to tasks in hierarchical task network planning  \cite{erol1996hierarchical}, where the tasks are processes inferred from the agent's execution (e.g., ``explore in this direction'') rather than goals, which need to be validated in a particular state (e.g., ``reach {coordinate} $(3,5)$''). {Second, the nature of our problem motivates a single unified reward for the meta-layer and action-layer rather than separate rewards. As already mentioned, this reward is the output of the classification layer. Lastly, the partial observability of our problem motivates derivation and use of policy-gradient approaches for learning the model parameters. As illustrated in \cite{mousavi2019multi}, such generalized policy gradient algorithms allow co-design of goal generator, action planner, and classifier modules. }
	
		Our methodology incorporates goal reasoning capabilities with deep reinforcement learning procedures for robot navigation by introducing intermediate goals, instead of requiring the robot to take a sequence of actions. In this way, our architecture provides transparency in terms of what the robot is trying to accomplish and, thereby, provides an explanation for its own course of action. {The statement of the classification problem is identical to that of  \cite{mousavi2019multi}, but with some important differences. In \cite{mousavi2019multi}, we employ multiple agents with a recurrent network architecture, while robots do not enjoy goal reasoning capabilities. }

\noindent{\emph{Related Literature:}} We cast the classification problem as a planning and perception mechanism with a three-layer architecture that is realized through a feedback loop. We are particularly interested in planning for perception. A related line of research is active perception: how to design a control architecture that learns  to complete a task and at the same time, to focus its attention   to collect necessary observations from the environment (see  \cite{whitehead1990active,aloimonos2013active}  and references therein). The coupling between action and perception has been also inspired by human body functionalities \cite{ballard1992hand}.  

Visual attention is another related line of work. It is based on the idea that for a given task, in general, only a subset of the environment may have necessary information, motivating the design of an attention mechanism \cite{tsotsos2011computational,balcarras2016attentional}. These have been motivating for saliency-based techniques for computer vision and machine learning, where the non-relevant parts of the data are purposely ignored \cite{mesquita2016object, bruce2015computational,bruno2019image,potapova2017survey,schauerte2016bottom}.  

\vspace{0.2cm}
	\noindent\emph{Notations:}  	 The $i$'th element of a vector $\pi$ is denoted by $\pi[i]$, where indexing may start from $0$.  For an integer $T >0$, $[T]$ denotes the sequence of labels  $[0,1,\dots,T-1]$. For two images $y_1 \in \R^{c_1 \times n \times n}$ and $y_2 \in \R^{c_2 \times n \times n}$ that have the same dimensions  but different number of channels, their concatenation is denoted by $\mathrm{concat}([y_1,y_2]) \in \R^{(c_1+c_2) \times n \times n}$. The categorical distribution over the elements of a probability matrix (or vector) $\pi$, whose elements add up to $1$, is denoted by $\mathrm{categorical}(\pi )$. For two probability vectors, $\pi_1,\pi_2 \in \R^D$, the cross-entropy between the corresponding categorical distributions is denoted by 
  	 $\mathrm{CrossEntropy}  (\pi_1,\pi_2)$. 
    
\section{Problem Statement} 

\begin{figure}[t]
	\begin{center}
\includegraphics[width=2.0cm]{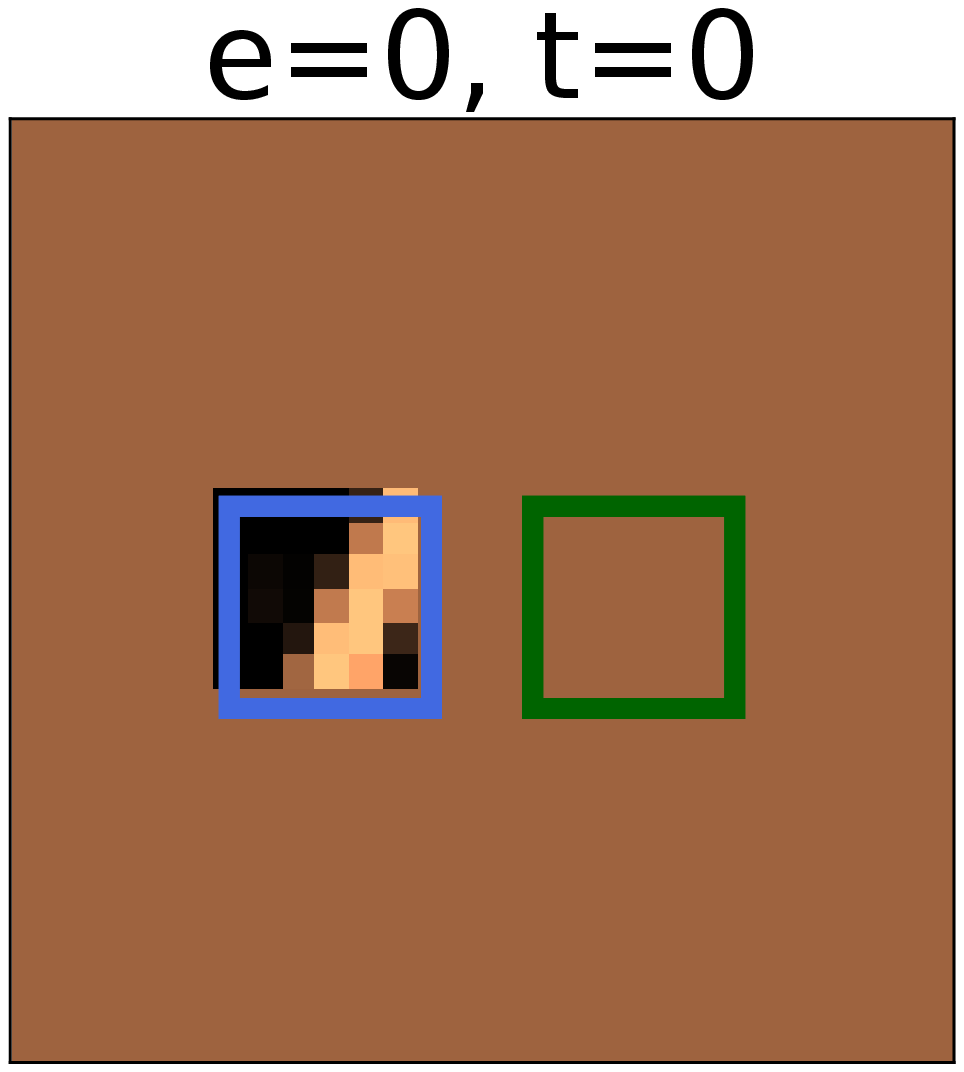}
\includegraphics[width=2.0cm]{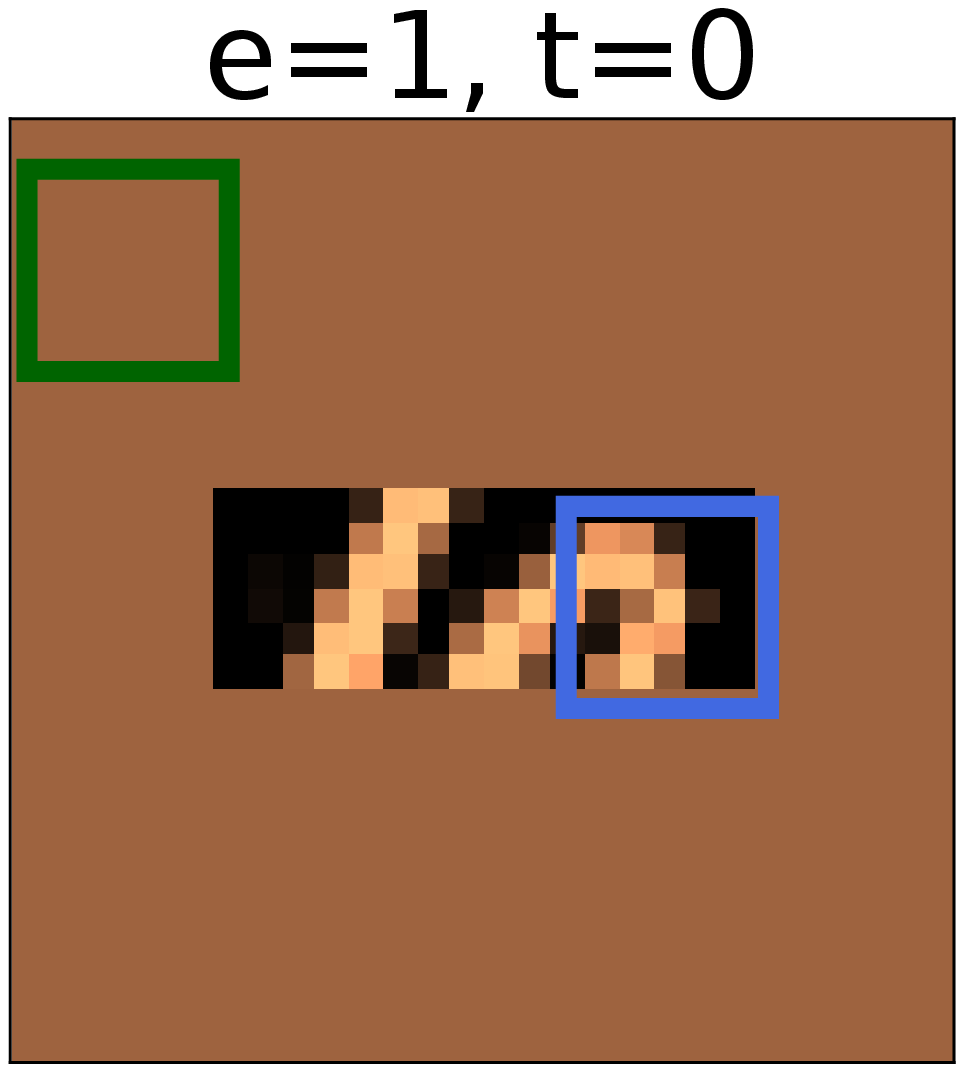}
\includegraphics[width=2.0cm]{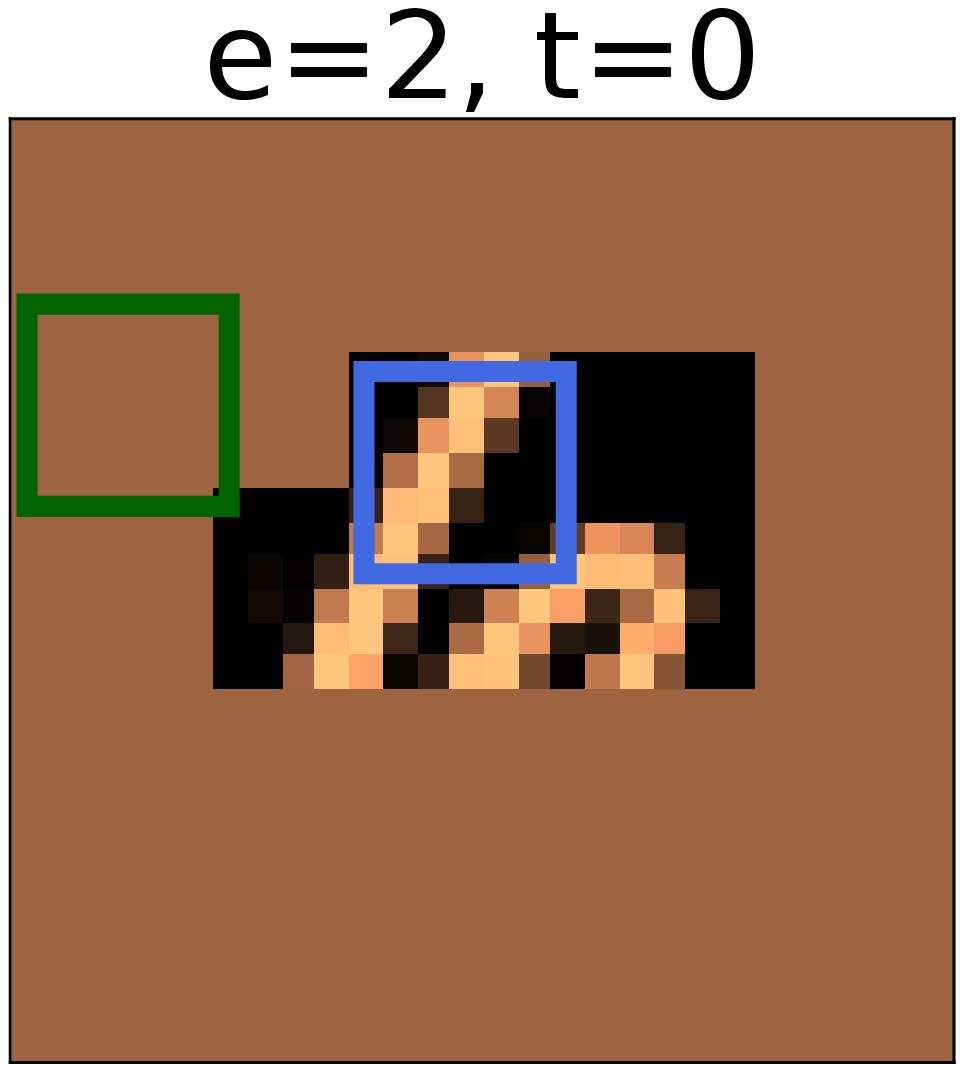}
\includegraphics[width=1.2cm]{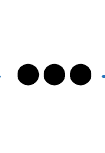}
\end{center}
$~~~~~~y(0,0)~~~~~$ $~~~~~y(1,0)~~$ $~~~~~~~~y(2,0)~~$
\caption{Snapshots of the proposed problem at the beginning of three episodes. The blue and green squares point to the current {position} of the agent and the goal of each episode. During each episode, the agent has moved towards the goal.  }
\label{fig:b}
\end{figure}

Let us consider an agent (robot) that is capable of moving in some pre-specified directions (such as up, down, right, and left) in order to explore an image (e.g., map of a region) during a sequence of $E > 0$ episodes, where the duration of each episode is $T > 0$ steps in time. 
For  integers $c, n >0$, we represent an instance of an image by a $c \times n \times n$ {matrix}. 
Suppose that at the beginning of episode $e \in [E]$ a goal $g(e)$ is assigned to the robot and at every time step $t \in [T]$ (within that episode), the robot moves towards $g(e)$ to discover a portion of image $x \in \R^{ c \times n \times n}$ based on its current pose $p(e,t) \in \R^2$. The robot takes an action to update its position.  Based on its past history, the agent has uncovered portions of $x$ up to time $t$, which is denoted by $y(e,t) \in \R^{c \times n\times n}$. The undiscovered portions of $x$ in $y(e,t)$ are set to $0$. Fig. \ref{fig:b} illustrates this scenario  through an example, where the discovered image $y(e,t)$, the robot's position, and its goal are demonstrated at different episodes and times.   
The {\it problem} is to design a layered  architecture that  generates meaningful goals and plans navigation towards assigned goals, with the objective of  performing image classifcation.

\section{A Multi-layered Architecture}

We propose an architecture where a robot collects local observations from an image, generates intermediate goals based on what it has been observed, takes local actions to move towards these goals, and, finally, makes a prediction based on the discovered information by the end of the last episode to classify the underlying image.  This architecture consists of three layers, where each receives a different set of information as their inputs. These inputs are  defined using some auxiliary internal variables. For given $e \in [E]$ and $t\in [T]$, we define an auxiliary image $l(e,t) \in \R^{n \times n}$ whose pixels are set to $1$ everywhere except over a $m \times m$ patch of pixels with $0$ values, {where $m$ denotes the width and height of the partial observation by the agent} This variable solely depends on the robot position $p(e,t)$. Similarly, we define an auxiliary image $h(e,t)\in \R^{n \times n}$ where the value of a pixel is set to $0$ if robot has visited that pixel before, otherwise to $1$. This variable keeps track of the history of the agent. 
  
  \begin{figure*}
  	\begin{center}
  	\includegraphics[width=14.1cm]{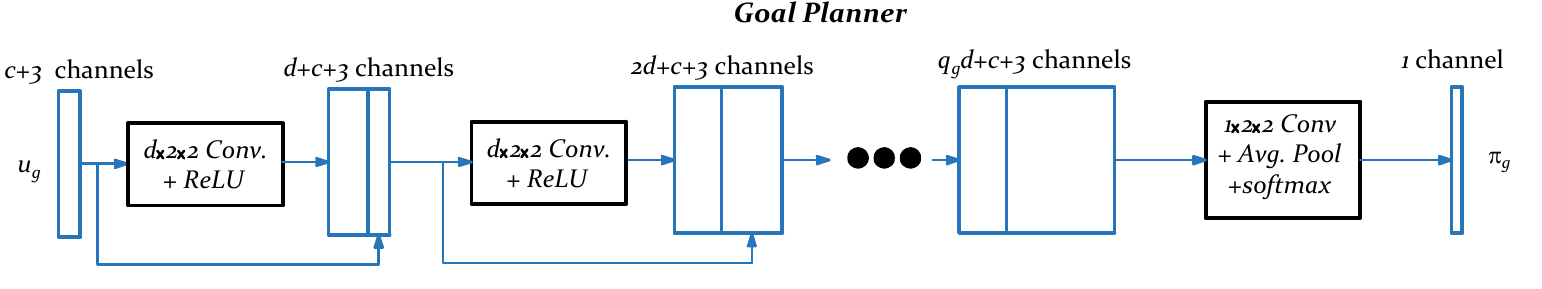}
  	\includegraphics[width=15.5cm]{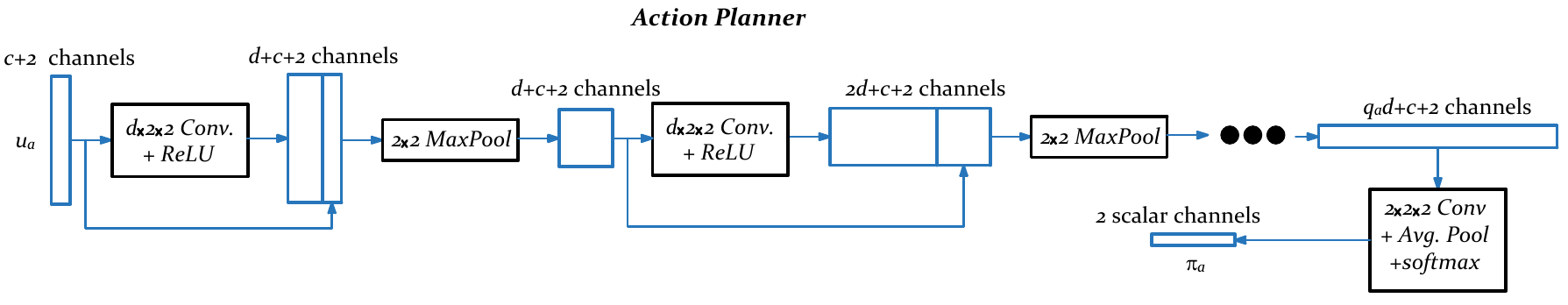}
  	\includegraphics[width=15.5cm]{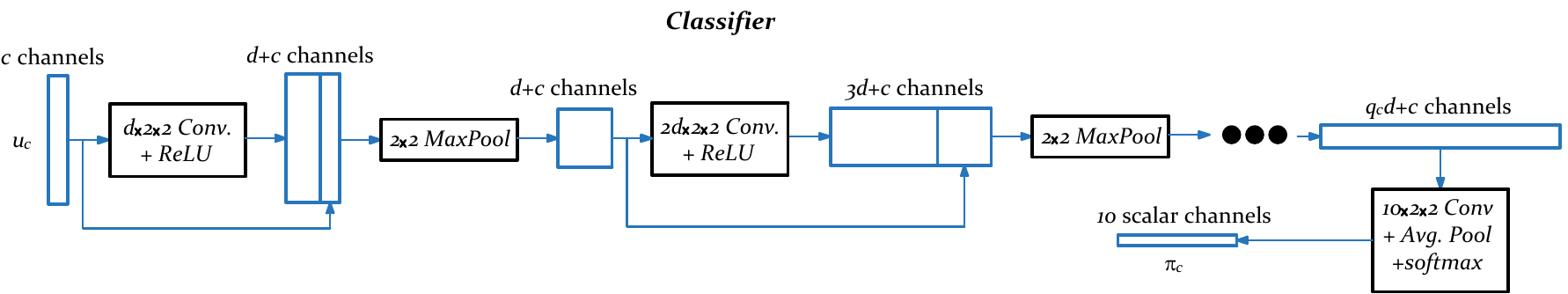}
  	\end{center}
  \caption{A schematic diagram of the 3-layered deep learning architecture for goal generator, action planner, and classifier. The dots correspond to repeating the preceding modules for $r$ times. In the planners, the number of channels in the convolutional filters is fixed and equal to $d$ in the consecutive layers. For the classification module, the number of output channels from the convolutions is doubled each time. Thus, in each case we will have {different numbers} of intermediate channels $q_g$, $q_a$, and $q_c$ (the components are not drawn). }
  \label{fig:schematic}
  \end{figure*}

\subsection{Goal Planner}

We consider a fully-convolutional architecture of ResNet style \cite{he2016deep} for the planner, where the skip connections are modified to have  concatenation form instead of summation (similar to densely connected architecture \cite{huang2017densely}). The top portion of Fig. \ref{fig:schematic} illustrates our architecture.  At the beginning of episode $e$,  information input $u_g(e)\in {\R^{(c+2) \times n \times n }}$ is formed by concatenating three inputs: 

\vspace{0.1cm}
\noindent {\it (i)}  Undiscovered image up to {this episode and instant}, which is defined  by
\begin{align}
y(e-1):=y(e-1,T-1)  \in \R^{c\times n \times n}. 
\end{align}
We recap that $y(e,t)$ is the undiscovered portions of the underlying image at episode $e$ and time $t$.

\noindent{\it(ii)} An image that encapsulates the  position of the robot in the environment by the end of the previous episode, which is defined by
\begin{align}
l(e-1):=l(e-1,T-1) \in \R^{n \times n}. 
\end{align}

\noindent{\it(iii)} An image that encapsulates the history of all visited {positions} up to that  episode, which is defined by  
\begin{align}
h(e-1):=h(e-1,T-1) \in \R^{n \times n}. 
\end{align}
We feed the following input to the planner    
\begin{align*}
u_g(e):=
\mathrm{concat}  (\,\left [\,y(e-1),l(e-1),h(e-1),g_l(e-1)\,\right ]\, ), 
\end{align*}
where $g_l(e-1)$ is derived from the  previous goal $g(e-1)$ according to a procedure that is explained at the end of this subsection. Then, we utilize the convolution architecture that outputs    a single channel $n \times n$ image. By applying $\mathrm{softmax}$ on this image, we arrive at {an} $n \times n$ probability matrix that can be characterized by a nonlinear map
\begin{align}
 \pi_g(e)
 =f_1\big(u_g(e);\theta_1\big),
\end{align}
where $\theta_1$ is a trainable parameter.
  We define a categorical probability distribution over the pixels using $\pi_g(e)$, which will allow us to sample goal $g(e) \in \R^2$ from this distribution 
\begin{align}
g(e)\,\sim \, \mathrm{categorical}\big(\pi_g(e)\big).  
\end{align}
As a feedback signal for this layer and action-layer in the next episode, auxiliary variable $g_l(e) \in \R^{n\times n}$ is created, which is an image whose pixel values are set to $0$ only at the {$m \times m$ patch corresponding to the goal $g(e)$} and $1$ elsewhere (similar to $l(e)$).

\subsection{Action Planner for Local Navigation}

During each episode, the robot takes $T$ actions towards an assigned goal.  It is assumed that the actions taken by the robot are at most a fixed  number of pixels to the left, right, up, or down. Given the goal of the episode, one can inspect that there is always at most one horizontal action (either left or right) and one vertical action (either up or down) that we count as moving towards the goal. Therefore, given  current {position} $p(e,t)$  and  goal $g(e)$,  the problem of planning local actions can be formulated as finding a probability vector $\pi_a(e,t) \in \R^2$ that will allow the robot to choose between vertical and horizontal actions and move towards the goal. In situations where only one of these actions takes the robot closer to the goal, we do not use this distribution.  More precisely, robot's action protocol is given by  
$$
a(e,t)= \left \{ \begin{array}{l l} \hspace{-0.1cm} \text{vertical action} & \text{if } p(e,t)[0]= g(e)[0]\\
\hspace{-0.1cm} \text{horizontal action} & \text{else if } p(e,t)[1]= g(e)[1] \\
\hspace{-0.1cm} \text{sample from dist.} & \text{otherwise} \\ 
 \end{array}  \right .. 
$$
To evaluate the probability vector $\pi_a$, we   consider a similar fully-convolutional architecture  for choosing the local actions; we refer to the middle portion of Fig. \ref{fig:schematic}.  The input to this architecture lives in $\R^{(c+3)\times n \times n}$ and is defined by  
\begin{align}
u_a(e,t)=\mathrm{concat} \big (\left [\,y(e,t),\,l(e,t),\,h(e,t),\,g_l(e)\,\right ]\big ). 
\end{align}
The convolutional mapping results in an image with $2$ channels. Then, we use global average-pooling from this output, which is followed by $\mathrm{softmax}$ normalization to get a vector $\pi_a (e,t) \in \R^2$. By composing all these maps, we can obtain the following characterization    
\begin{align}
\pi_a(e,t)=f_2\big(u_a(e,t);\,\theta_2\big),
\end{align}
where $\theta_2$ is a trainable parameter. 
We construct a categorical  distribution which will enable the robot to select among vertical or horizontal actions via random sampling,  i.e.,
\begin{align}
a(e,t)\,\sim \, \mathrm{categorical}\big(\pi_a(e,t) \big).  
\end{align}

\subsection{Image Classifier} 

A similar convolutional architecture is considered for the classification module; we refer to the bottom portion of Fig. \ref{fig:schematic}.  Classification is conducted at the end of the last episode, i.e., {at episode $E-1$ and time step $T-1$}. Let us tag the last explored image  by 
$y_f:=y(E-1,T-1) \in \R^{c \times n \times n}.$
This will be the input to the classifier, i.e., 
\begin{align}
u_c = y_f. 
\end{align}
The output of the convolutional layer has $D$ channels, which is global average pooled before applying $\mathrm{softmax}$ to get the prediction vector $\pi_c \in \R^D$. Similar to the other two layers, the corresponding nonlinear map can be represented by
\begin{align}
\pi_c=f_3\big(u_c;\,\theta_3\big),
\end{align} where $\theta_3$ is a trainable parameter. The reward is defined as 
\begin{align}\label{eq:r}
r=-\mathrm{CrossEntropy} \Big (\,\pi_c\,,\,\pi_c^l \,\Big ),
\end{align} 
in which {$\pi_c^l \in \R^D$ is the label probability vector. This vector is equal to unit coordinate vector in $j$'th direction, where $j \in [D]$ is the label.}

\section{Reinforcement Learning Algorithm} 

We build upon our ideas from \cite{mousavi2019multi} and develop a learning algorithm to train various layers in our architecture. 
The robot's objective is to find an  unbiased estimator for the expected reward whenever the reward of the reinforcement learning explicitly depends on the parameters of the neural network. 	Let us put all trainable parameters in one vector and represent it by  
$
\Theta:=\left [\theta_1^T,\theta_2^T,\theta_3^T \right ]^T 
$. The set of all  trajectories is shown by  $\mathcal{T}$ and the corresponding reward to a given trajectory $\tau \in \mathcal{T}$  by $r^\tau$. The objective is to maximize the expected reward, i.e.,   
$$ \underset{\Theta}{\mathrm{maximize}}{}~J(\Theta),$$
where   
$ J(\Theta) =\mathbb{E} \{r^\tau\}  ={\sum_{\tau \in \mathcal{T}}  \pi^\tau \,  r^\tau }$ and $\pi^\tau$ is the probability of choosing goals and actions  given the value of the current parameter  $\Theta$.
The gradient of $J$ with respect to $\Theta$ can be written as 
\begin{align}
\nabla J\,=\,\sum_{\tau \in \mathcal{T}} r^\tau \nabla  \pi^\tau +  \pi^\tau \nabla r^\tau.
\end{align}
The REINFORCE algorithm
\cite{sutton2000policy}
helps us rewrite the first term using the following identity 
$
\nabla \pi^\tau=\pi^\tau  \nabla (\log \pi^\tau).   
$
Then, one can verify that  
\begin{align}\label{eq:nabla}
\nabla J &\,=\,\sum_{\tau \in \mathcal{T}} \pi^\tau \nabla (\log \pi^\tau)r^\tau + \pi^\tau \nabla r^\tau \\
& \notag \, =\, \mathbb{E}\{\nabla (\log \pi^\tau)r^\tau+\nabla r^\tau\}.
\end{align}
Suppose that $N$ independent trajectories are created, i.e., $N$ rollouts\footnote{A rollout is executing a fixed policy given an identical initial setting with a random seed. Different rollouts are required when the outcome of the game is uncertain (i.e., stochastic)\cite{sutton2018reinforcement}.}, where   
$\pi^{(k)}$ and $r^{(k)}$ denote the probability of   this particular trajectory and the resulting reward, respectively, for $k=1,\dots,N$.   Let us define  $\hat J$ to be 
\begin{align}\label{eq:hatJ}
\hat J\,:=\,\dfrac{1}{N} \Big ( \sum_{k=1}^{N} \log \pi^{(k)} r_d^{(k)}+r^{(k)} \Big ),
\end{align}
where the value of the quantity $r_d^{(k)}$ is $r^{(k)}$, while it has been detached from the gradients. {This means that a machine learning {algorithm} should treat $r_d^{(k)}$ as a non-differentiable scalar during training\footnote{The reason for this treatment is {because of} the idea behind the chain rule: in $(fg)'=f'g+g'f$, $f$ and $g$ in the right hand side correspond to being kept constant while the other term varies.} 
Then, we inspect that 
\begin{align}
\mathbb{E} \left \{\nabla \hat J \right \}\, =\, \nabla J,
\end{align}
i.e., $\nabla \hat J$ is an unbiased estimator of $\nabla J$ given by \eqref{eq:nabla}. 
This justifies the use of approximation  
$
\nabla J \, \approx \, \nabla \hat J. 
$

\begin{remark}
	The first term inside the summation in \eqref{eq:hatJ} is identical to the quantity that is derived in the policy gradient method with a reward which is independent of the parameters, i.e., the REINFORCE algorithm \cite{sutton2000policy}.  The second term indicates that reward  directly depends on  $\Theta$.  For example, if all goals and actions have equal probability of being selected, then it will suffice to consider {only} the second term inside the summation in \eqref{eq:hatJ}. 
\end{remark} 

\subsection{Hierarchical Training} \label{subsec:hi}

\begin{table}[]
	\resizebox{8.5cm}{!}{%
		\begin{tabular}{|l|c|c|c|}
			\hline 
			Layer &  being trained & trained \&  fixed & i.i.d.  \\ \hline\hline
			Classifier & \checkmark  & \checkmark  & $\times$  \\ \hline
			Goal Planner & \checkmark & \checkmark & \checkmark  \\ \hline
			Action Planner & \checkmark &  \checkmark & \checkmark \\ \hline
		\end{tabular}%
	}
\caption{Different possibilities for training of different layers. } 
\label{table:modes}
\end{table}

 \begin{figure*}[t]
	\begin{center}
		
		\includegraphics[width=2.1cm]{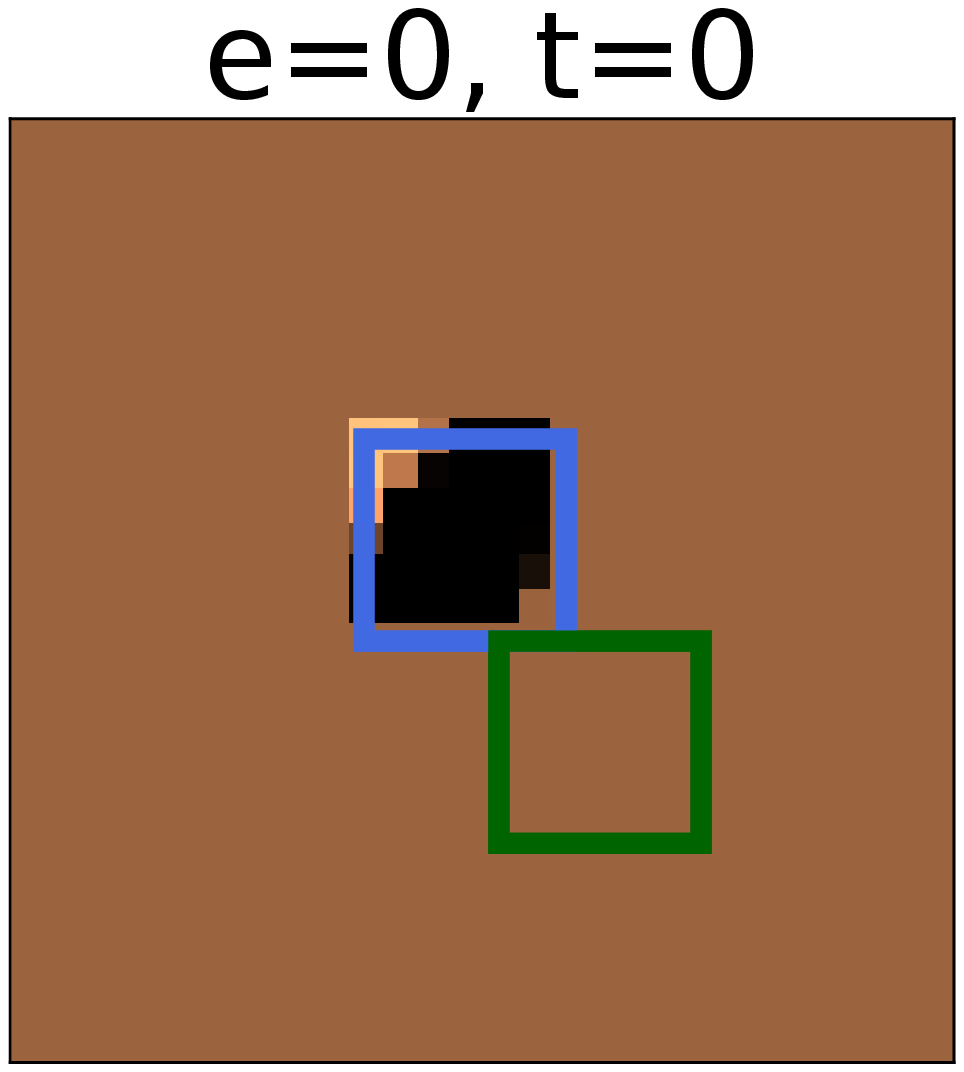} 
		\includegraphics[width=2.1cm]{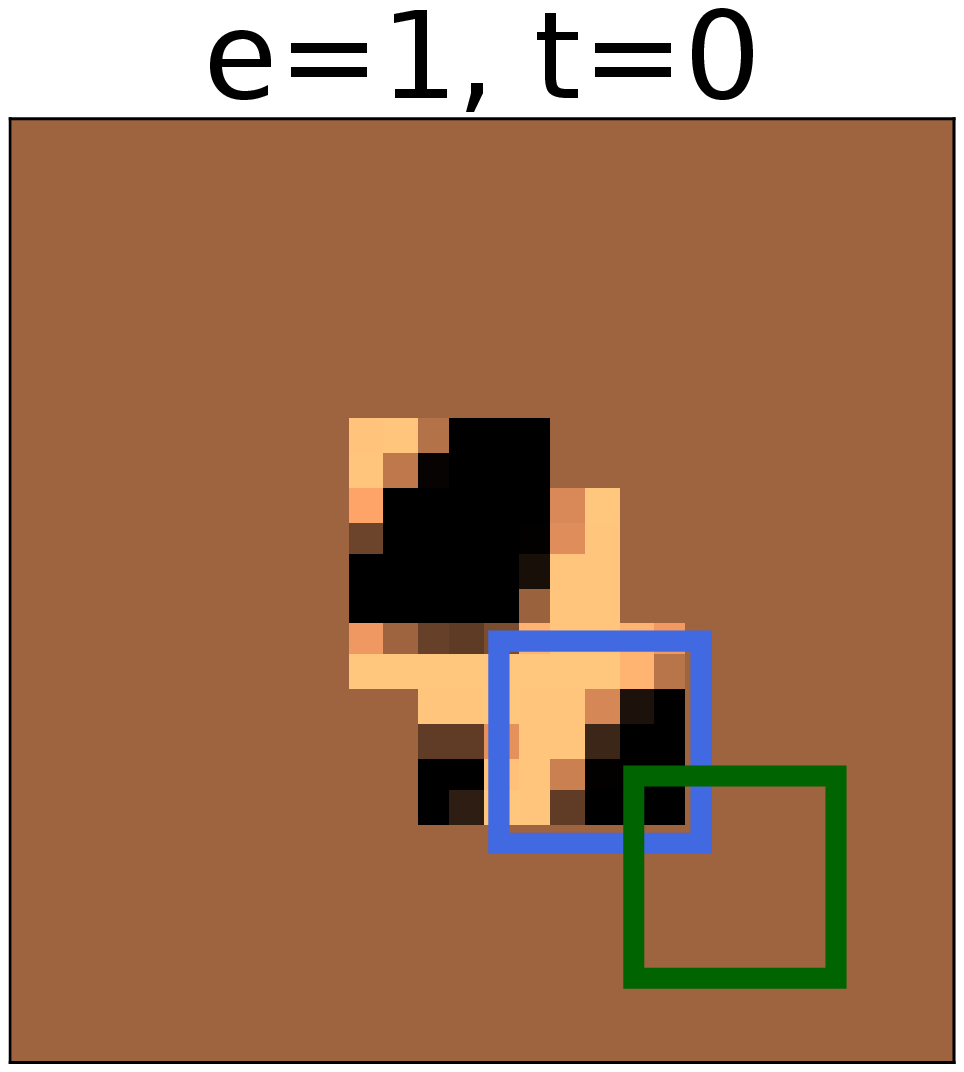}
		\includegraphics[width=2.1cm]{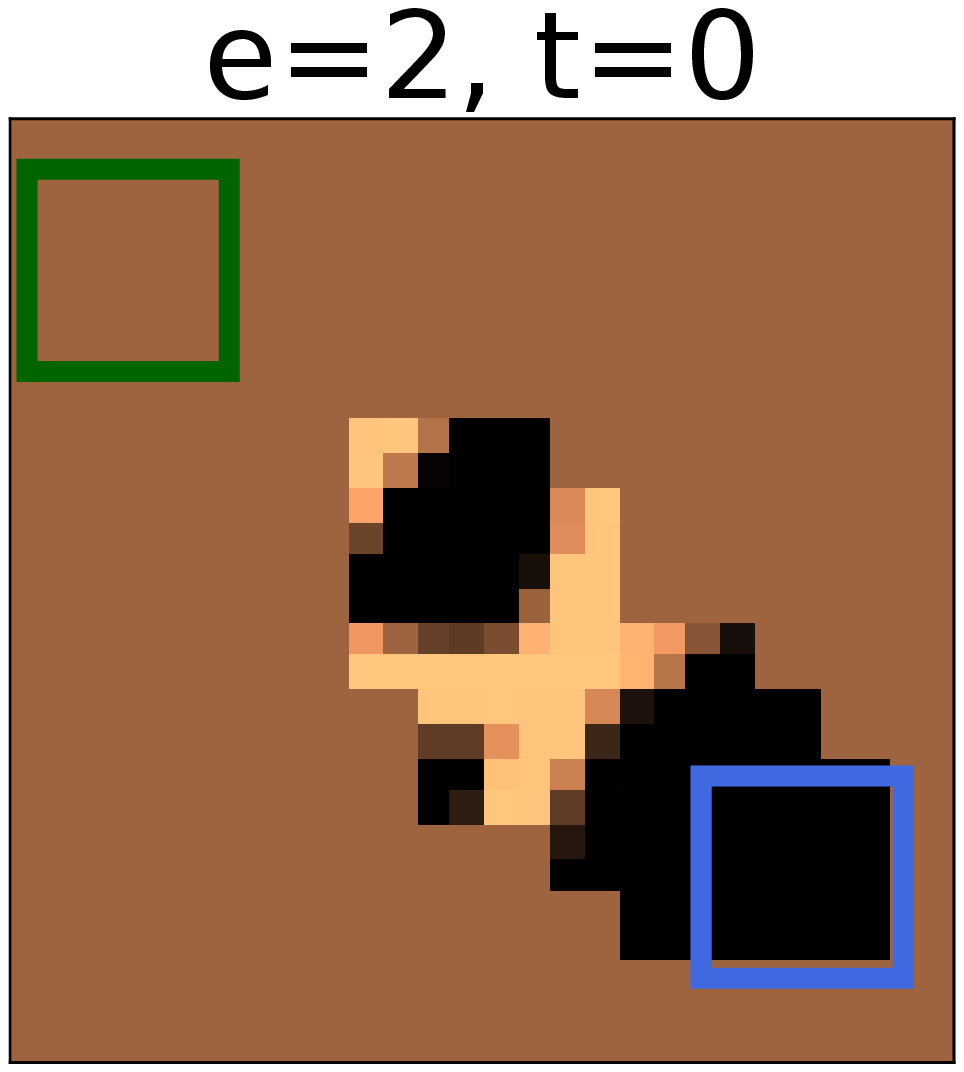}
		\includegraphics[width=2.1cm]{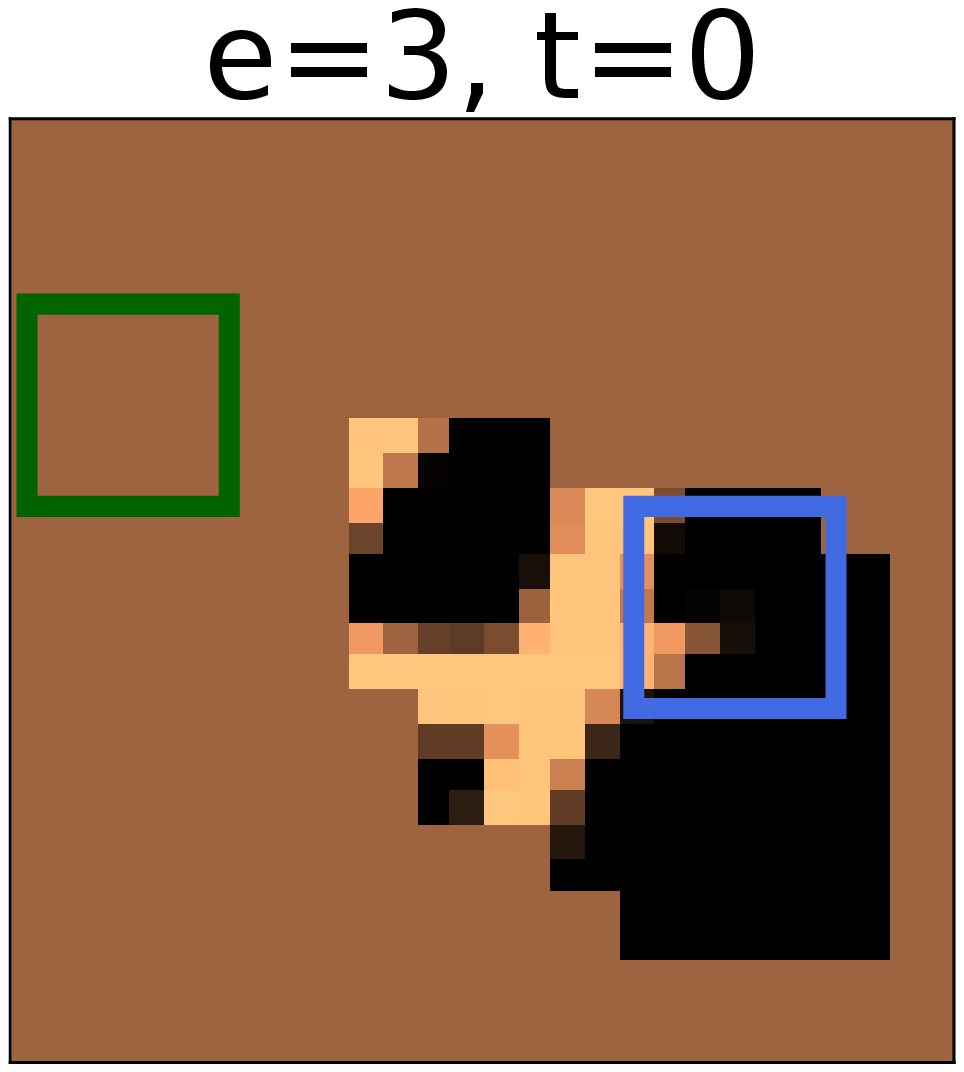}
		\includegraphics[width=2.1cm]{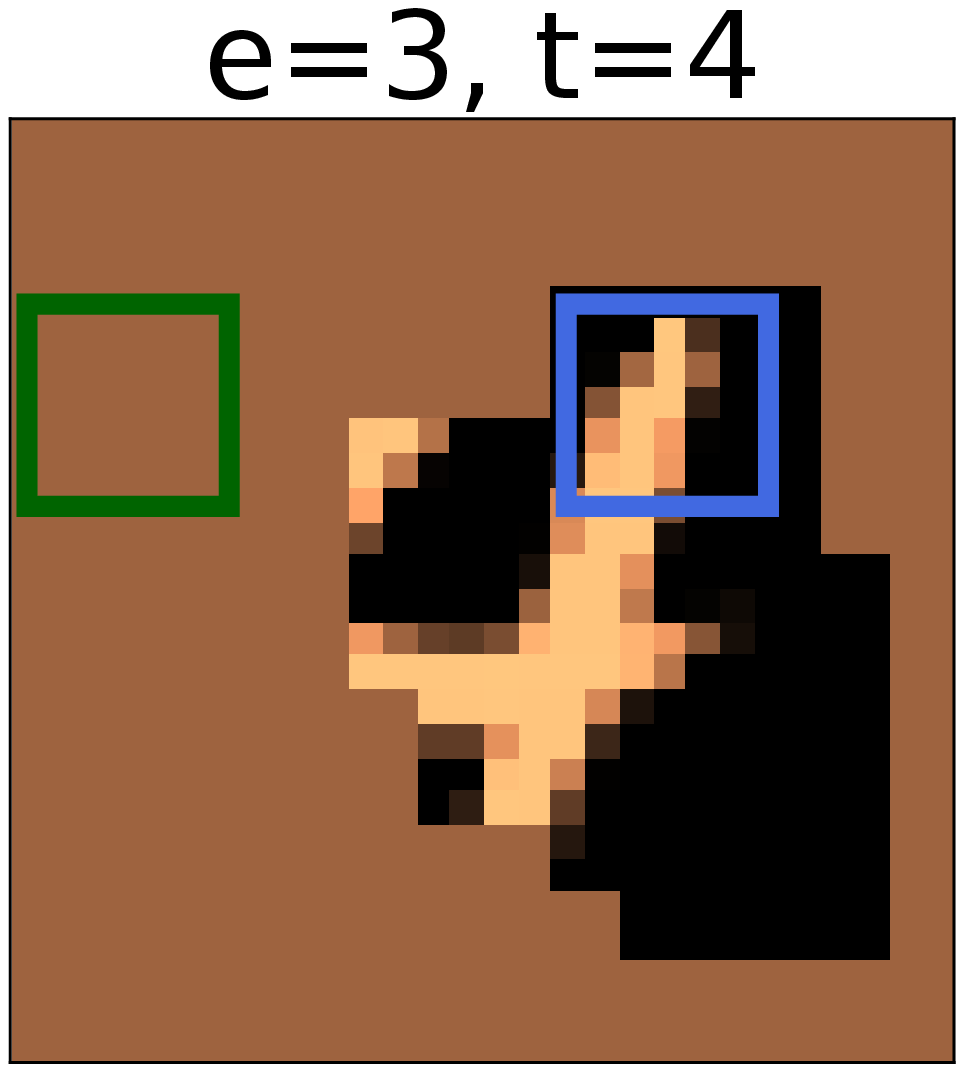}
		\includegraphics[width=2.1cm]{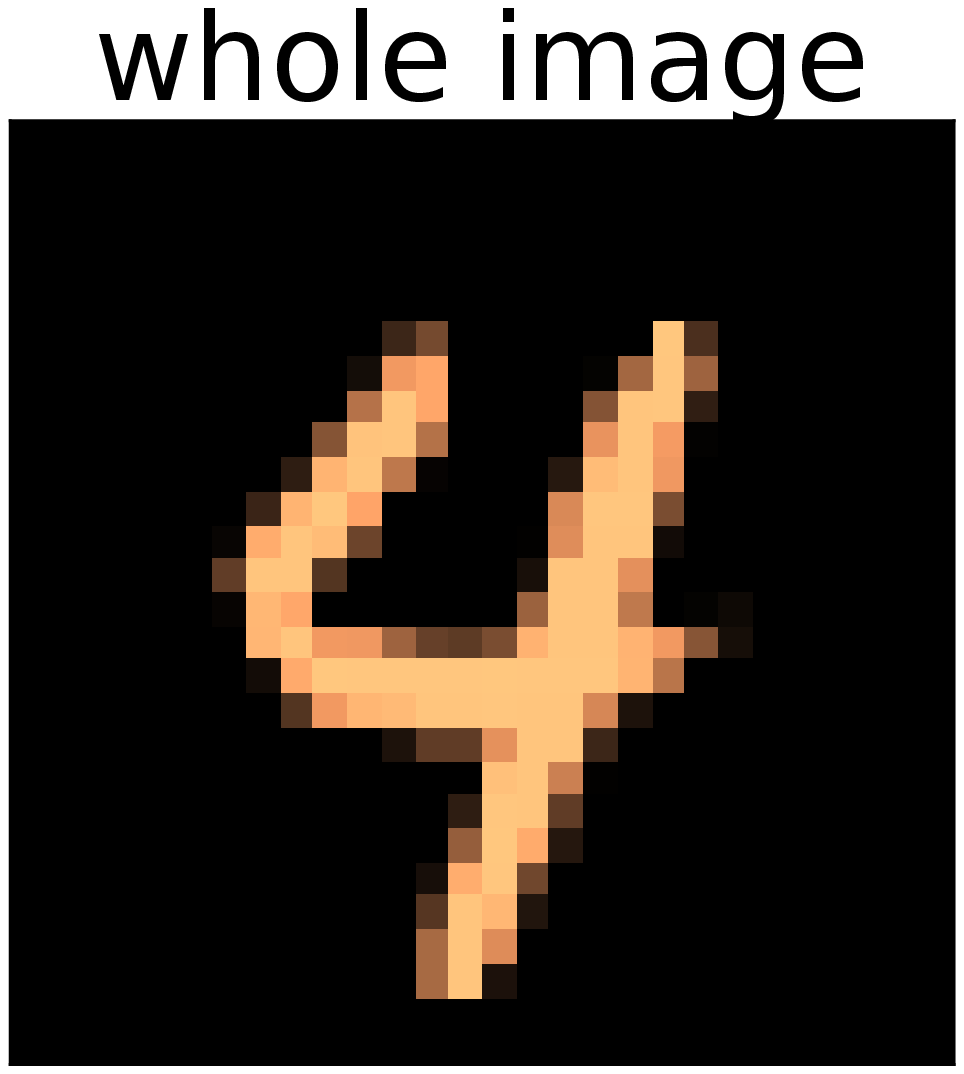}
		\includegraphics[width=2.5cm]{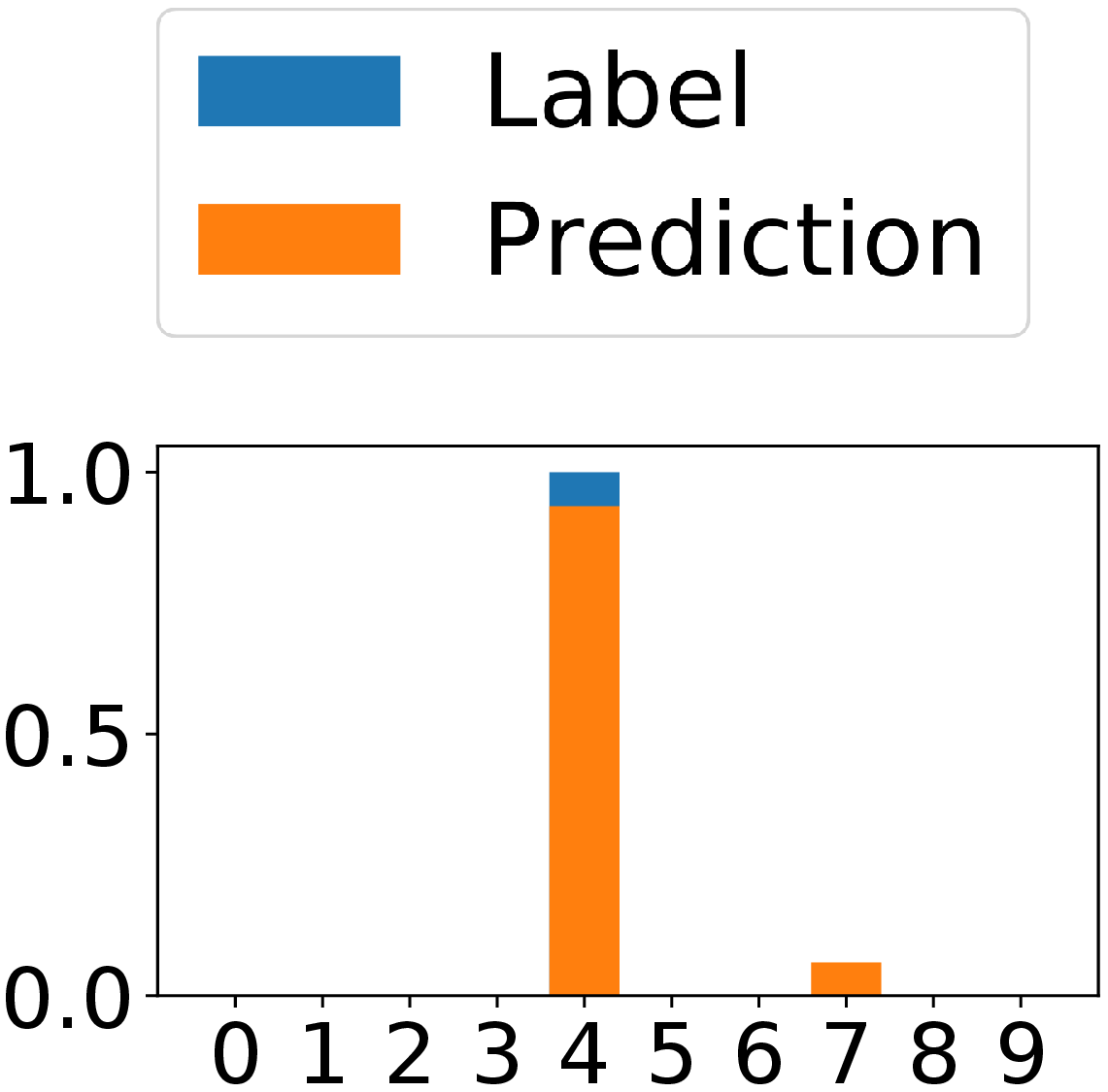}

		\includegraphics[width=2.1cm]{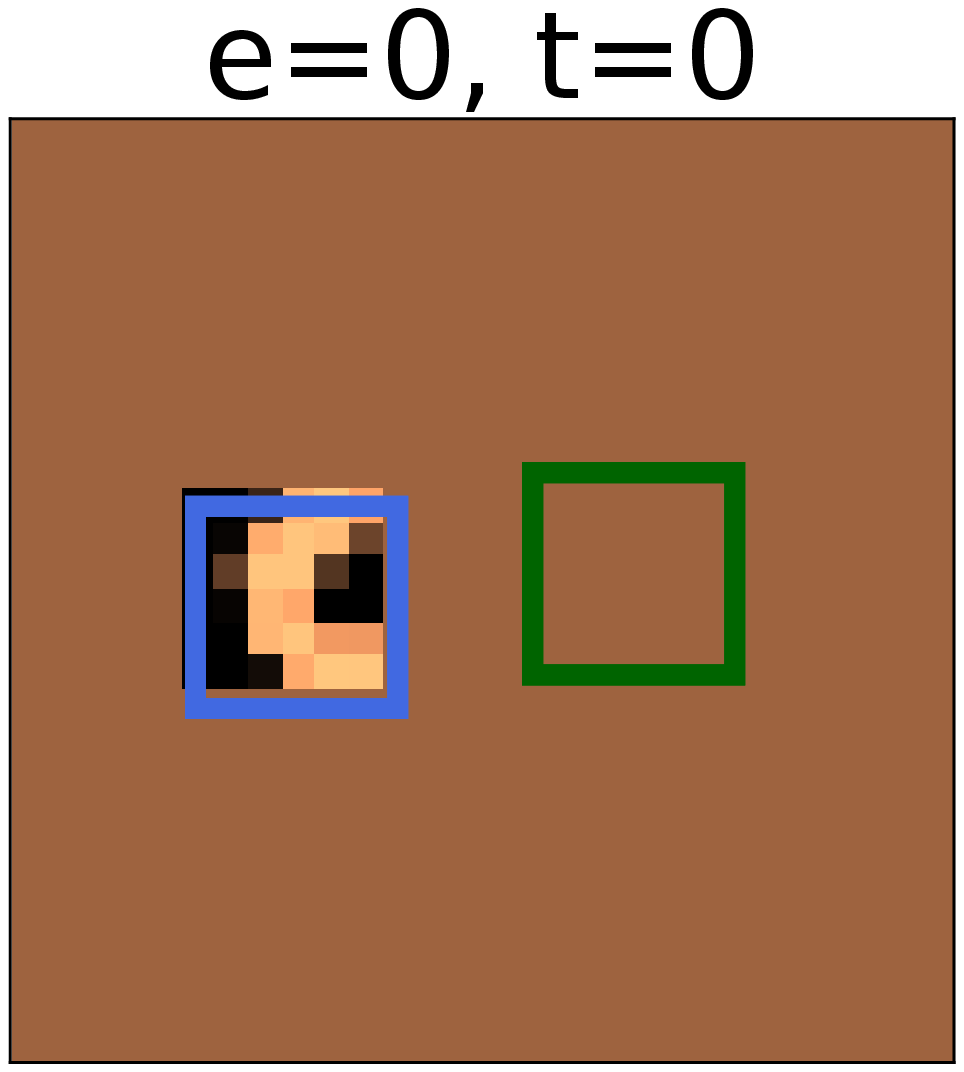} 
		\includegraphics[width=2.1cm]{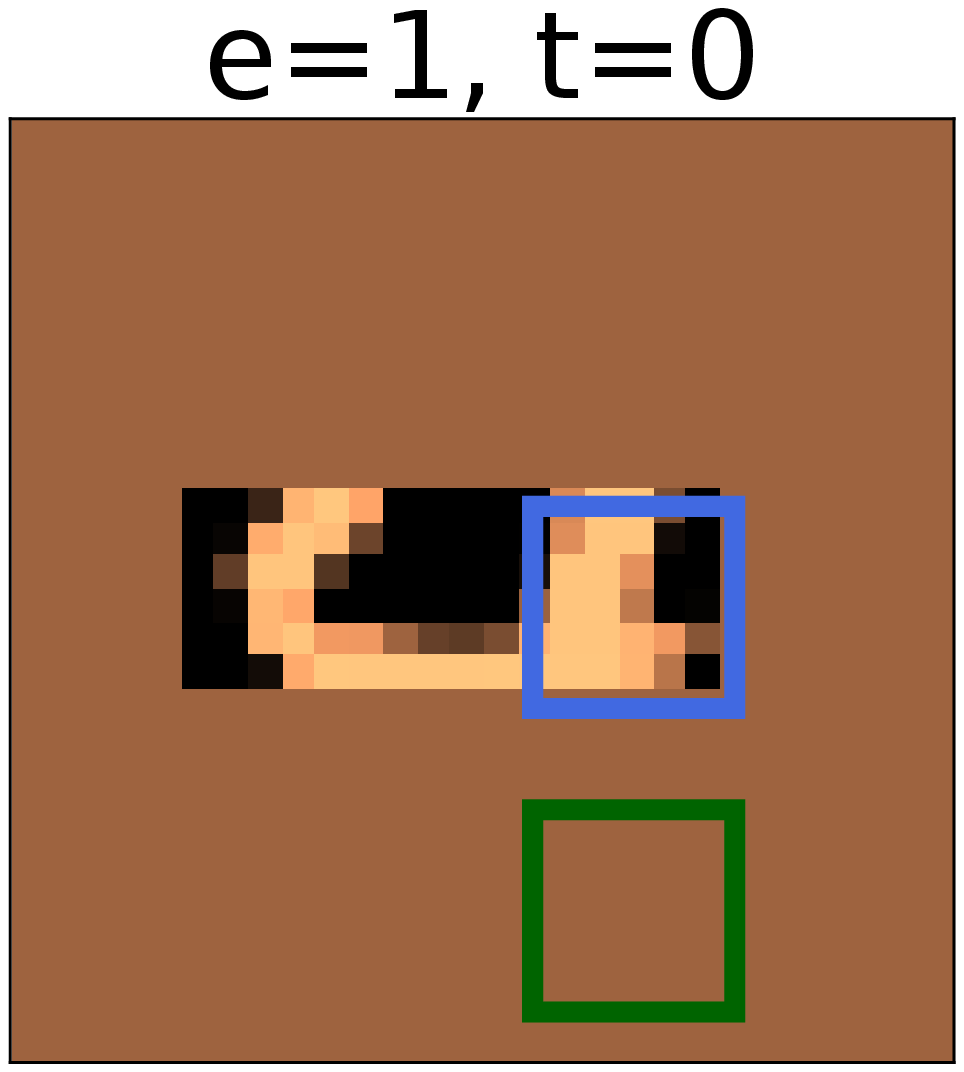}
		\includegraphics[width=2.1cm]{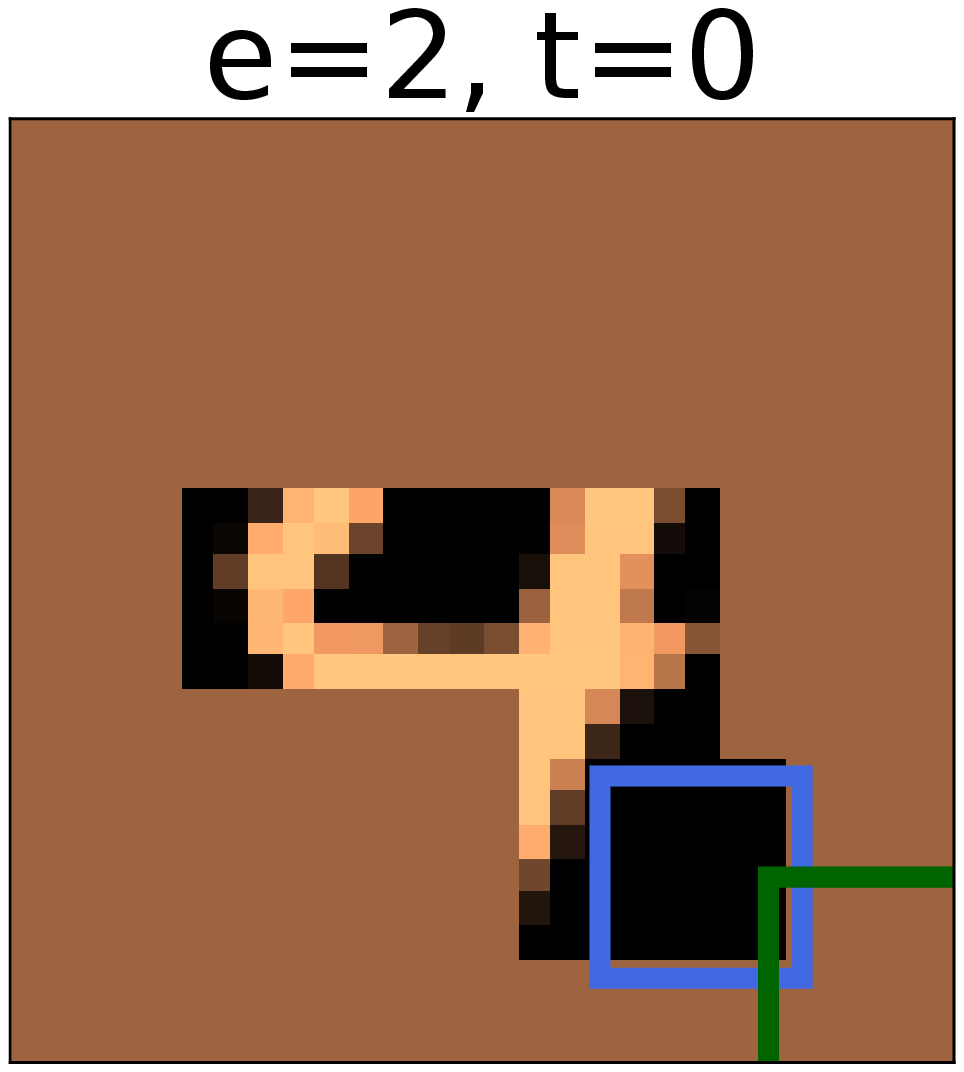}
		\includegraphics[width=2.1cm]{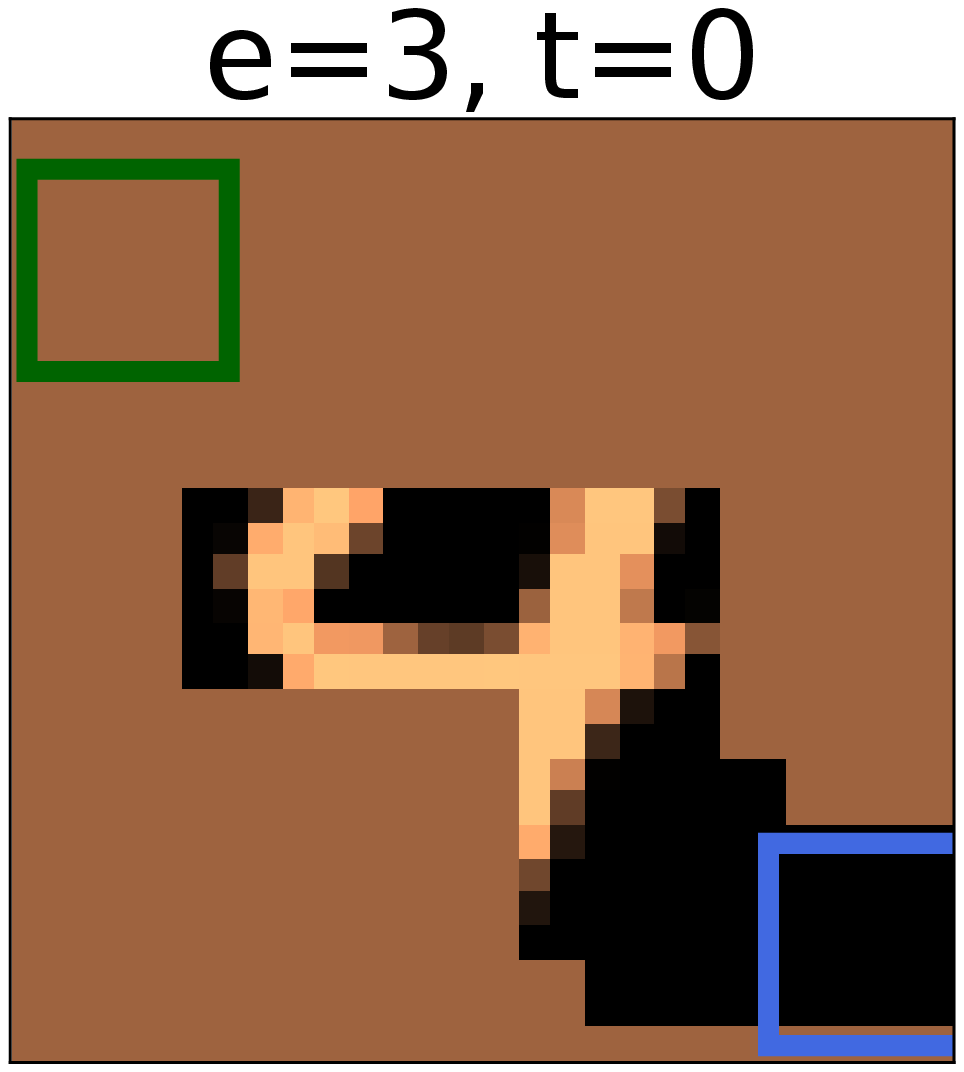}
		\includegraphics[width=2.1cm]{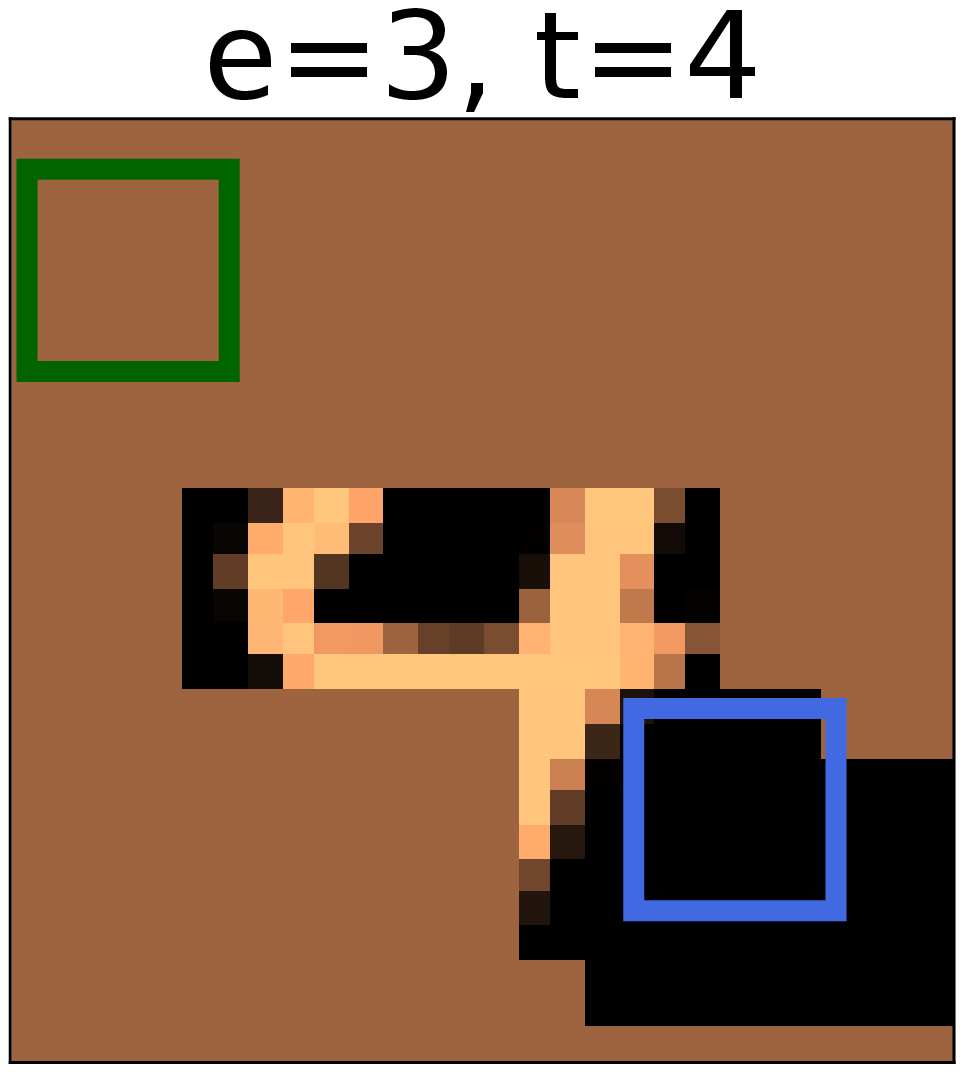}
		\includegraphics[width=2.1cm]{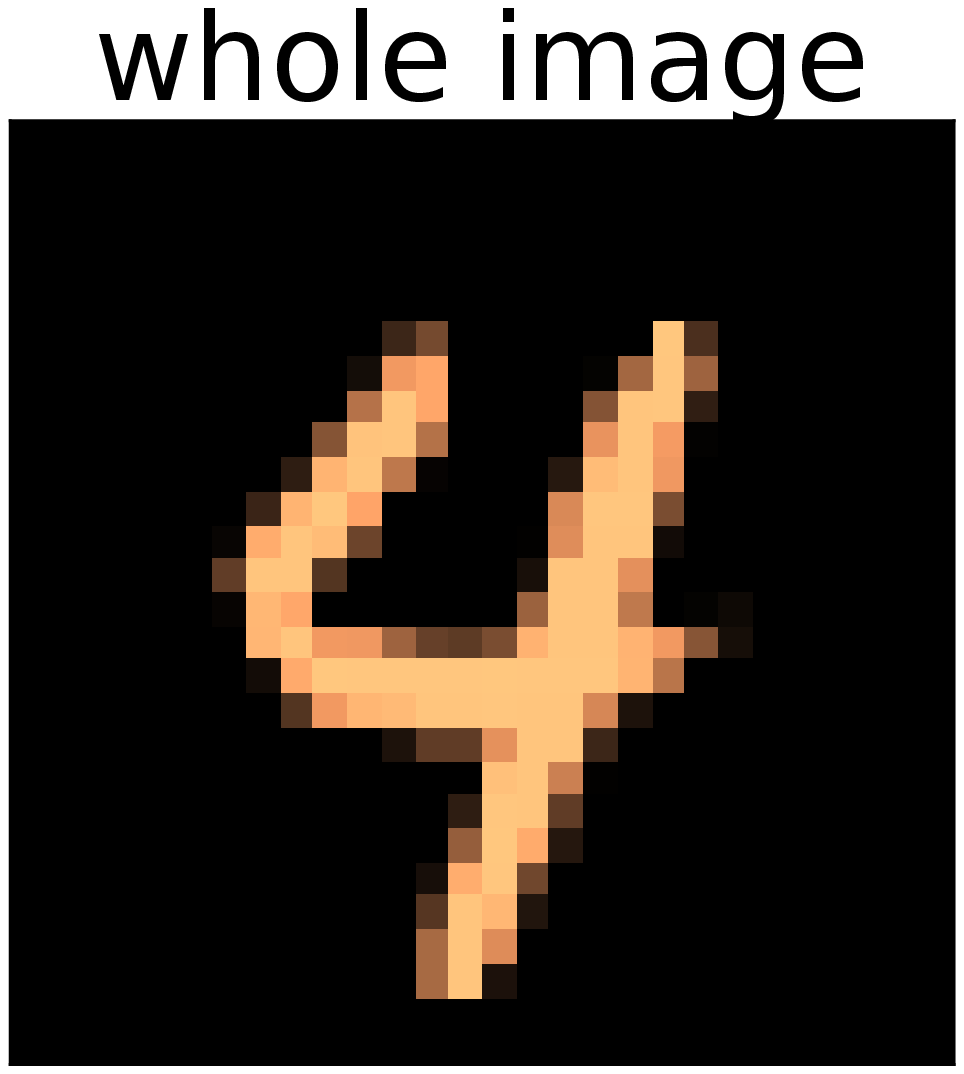}
		\includegraphics[width=2.5cm]{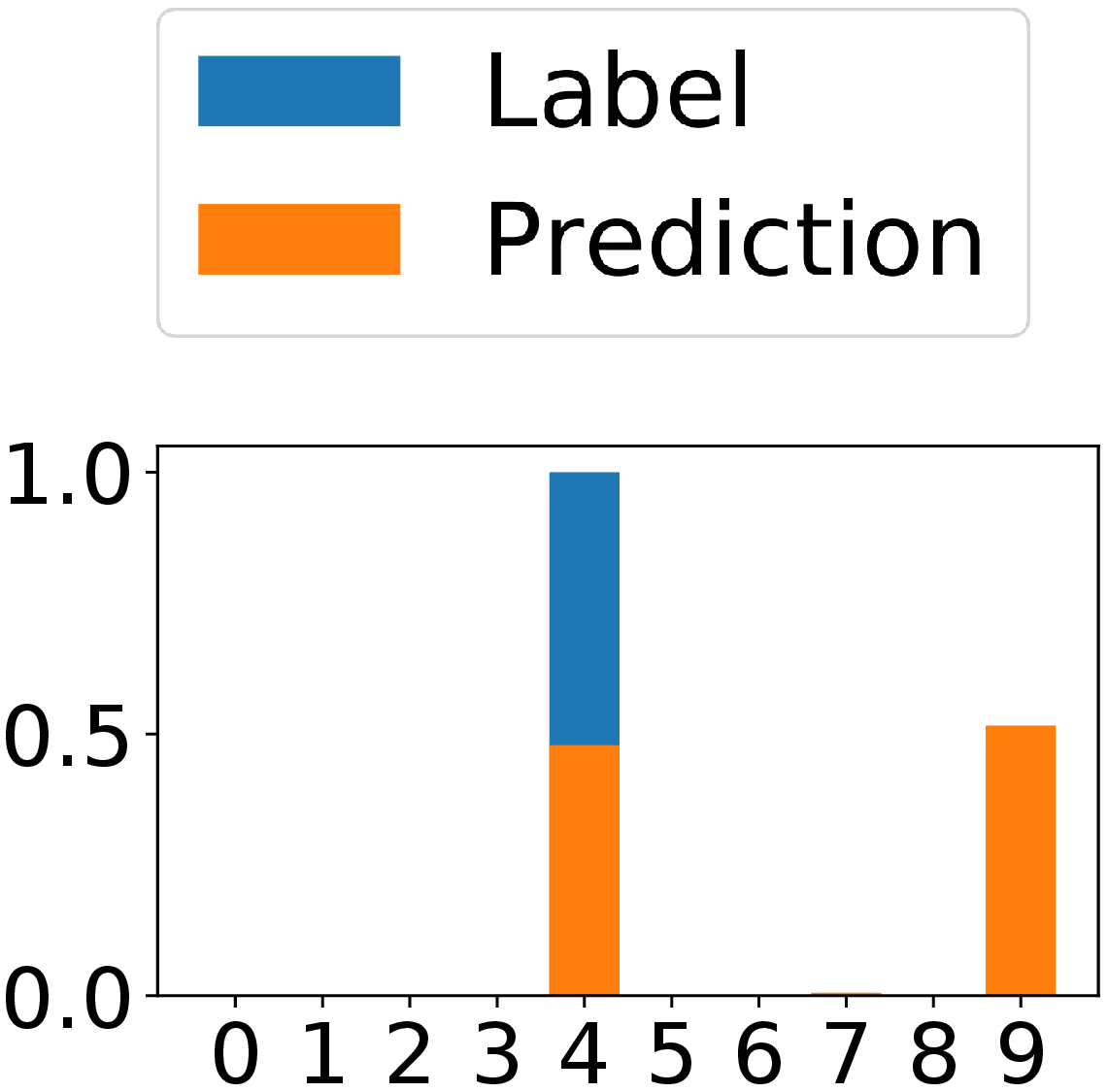}

		\includegraphics[width=2.1cm]{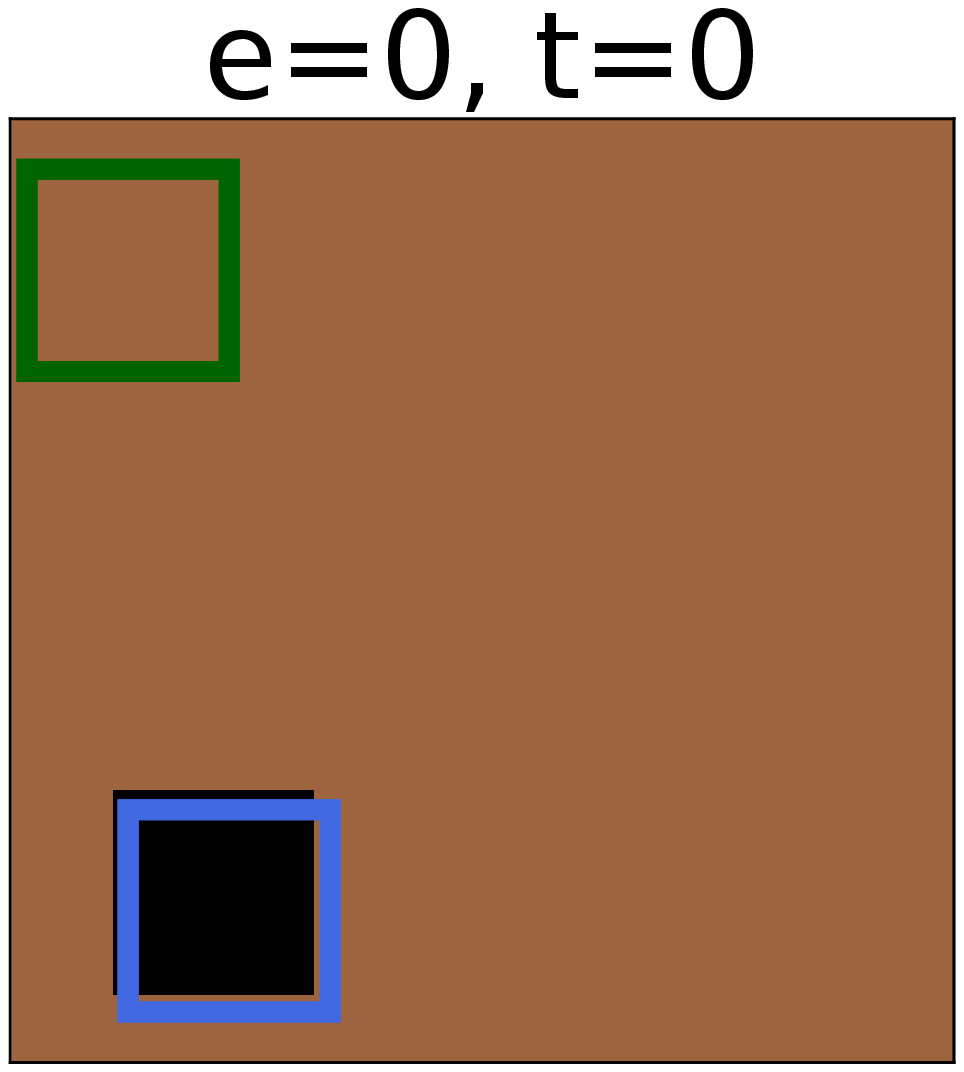}
		\includegraphics[width=2.1cm]{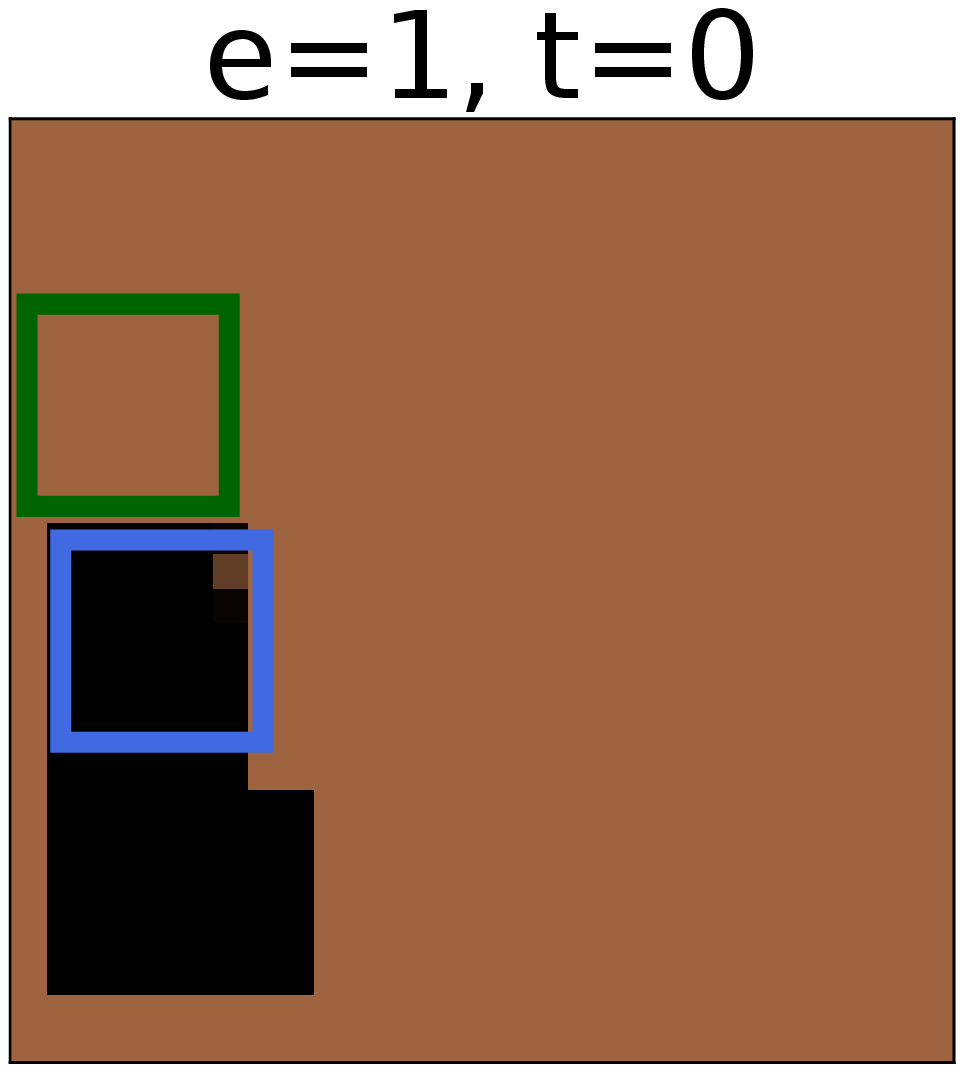}
		\includegraphics[width=2.1cm]{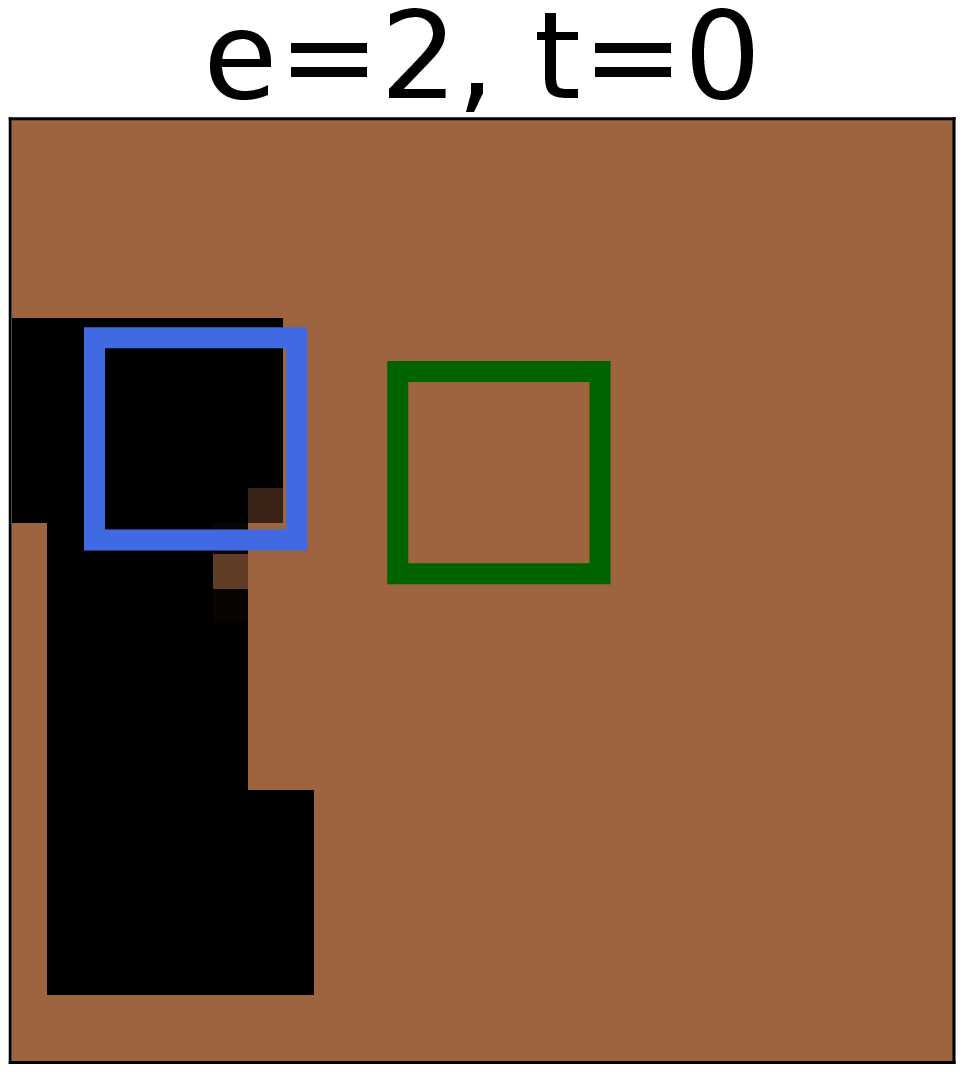}
		\includegraphics[width=2.1cm]{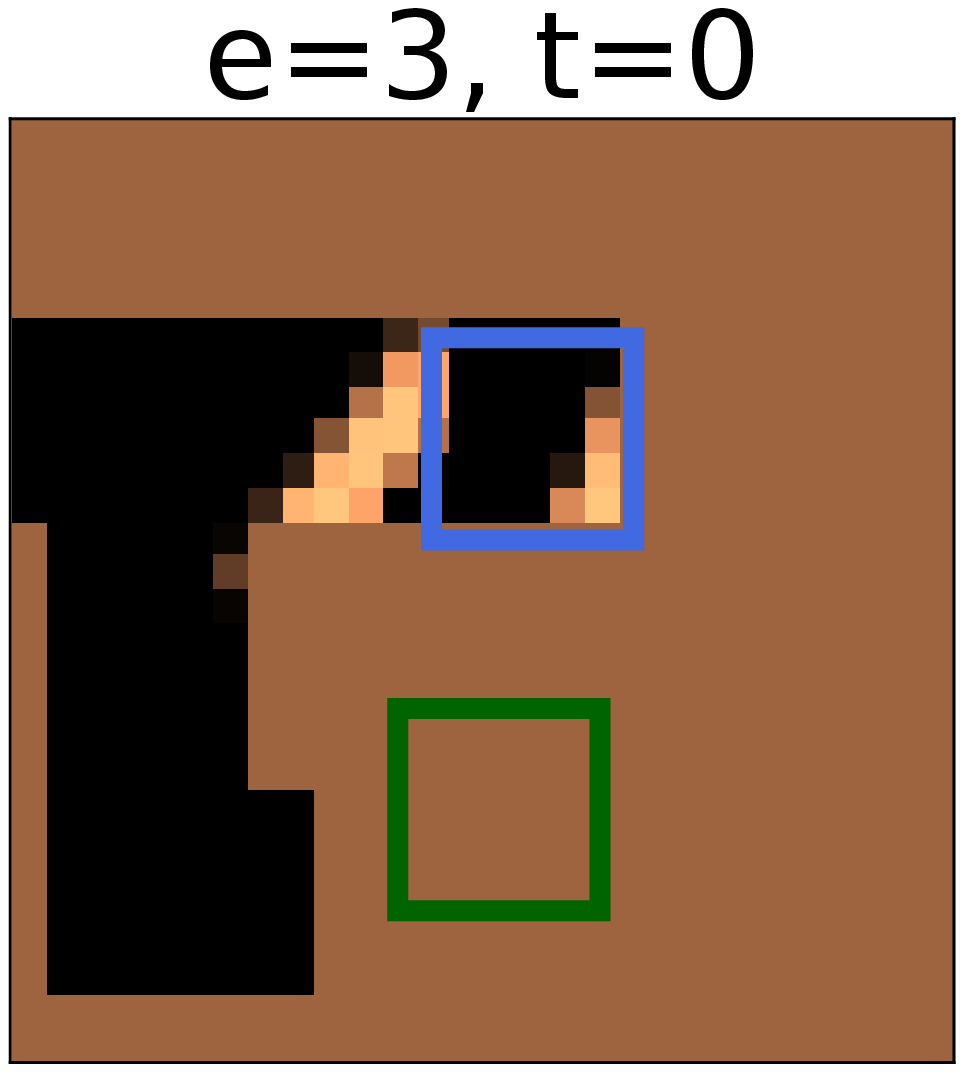}
		\includegraphics[width=2.1cm]{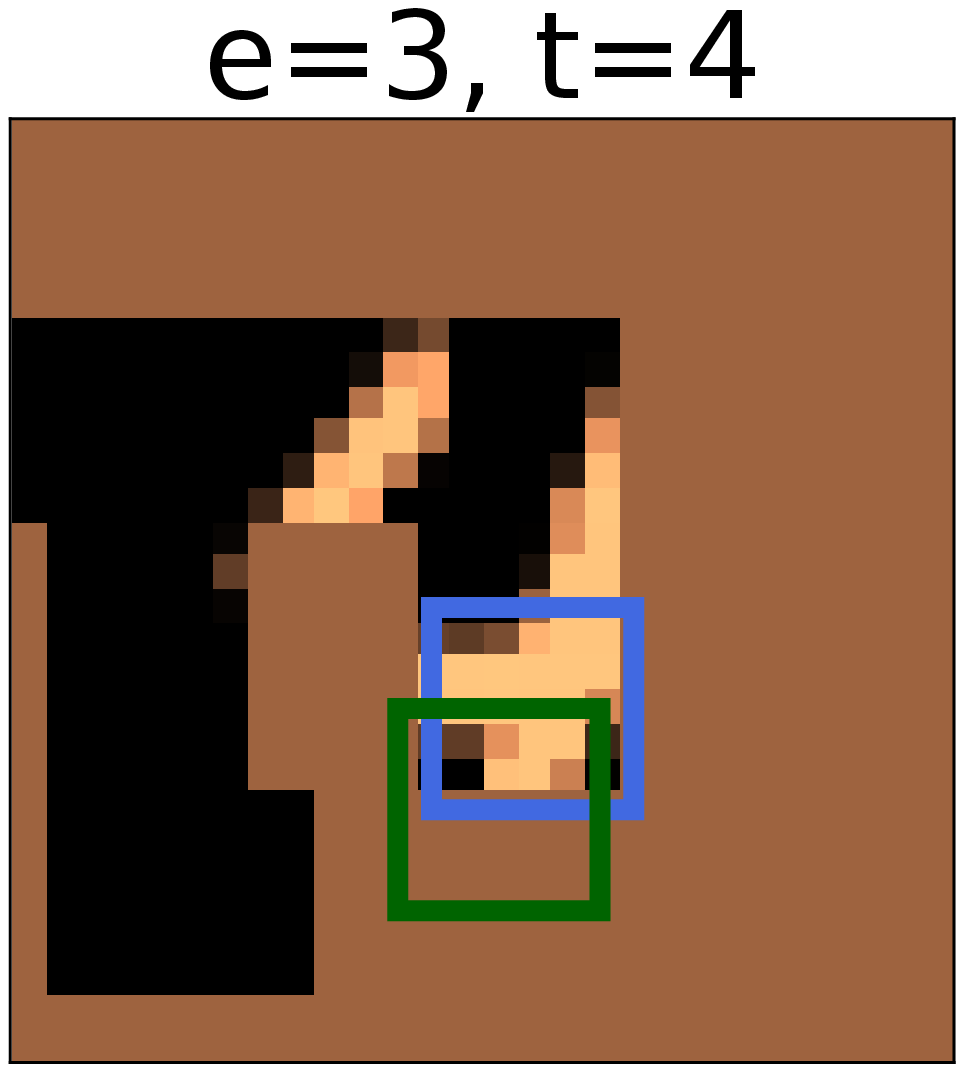}
		\includegraphics[width=2.1cm]{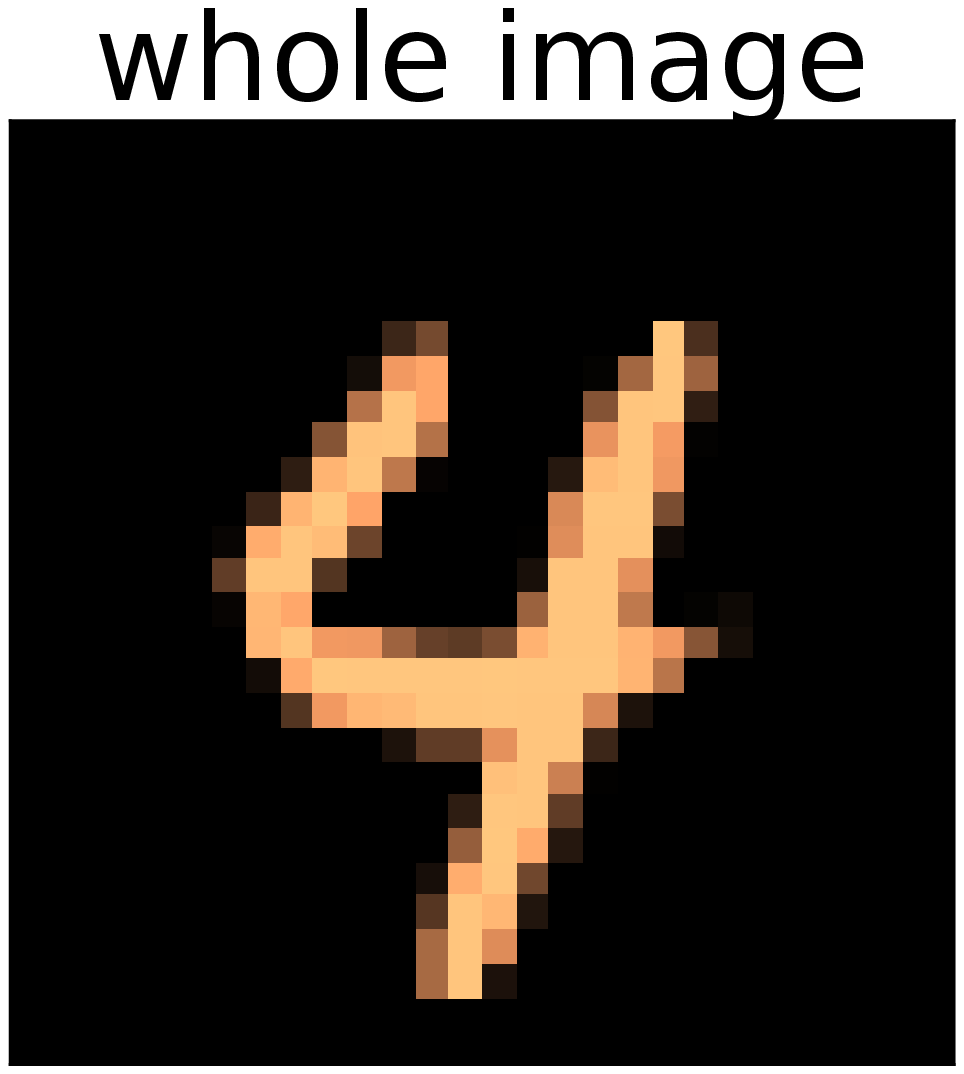}
		\includegraphics[width=2.5cm]{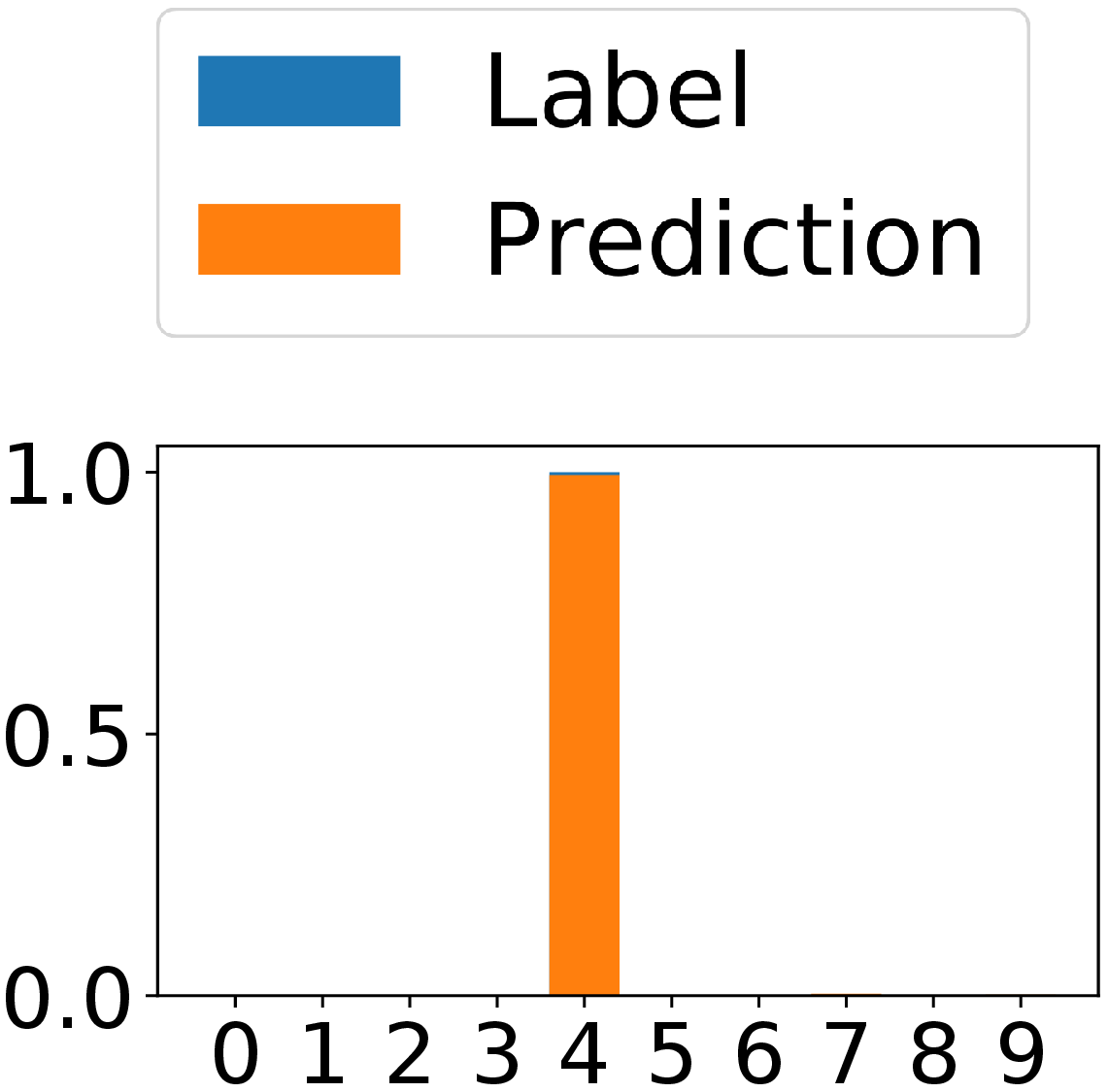}

		\includegraphics[width=2.1cm]{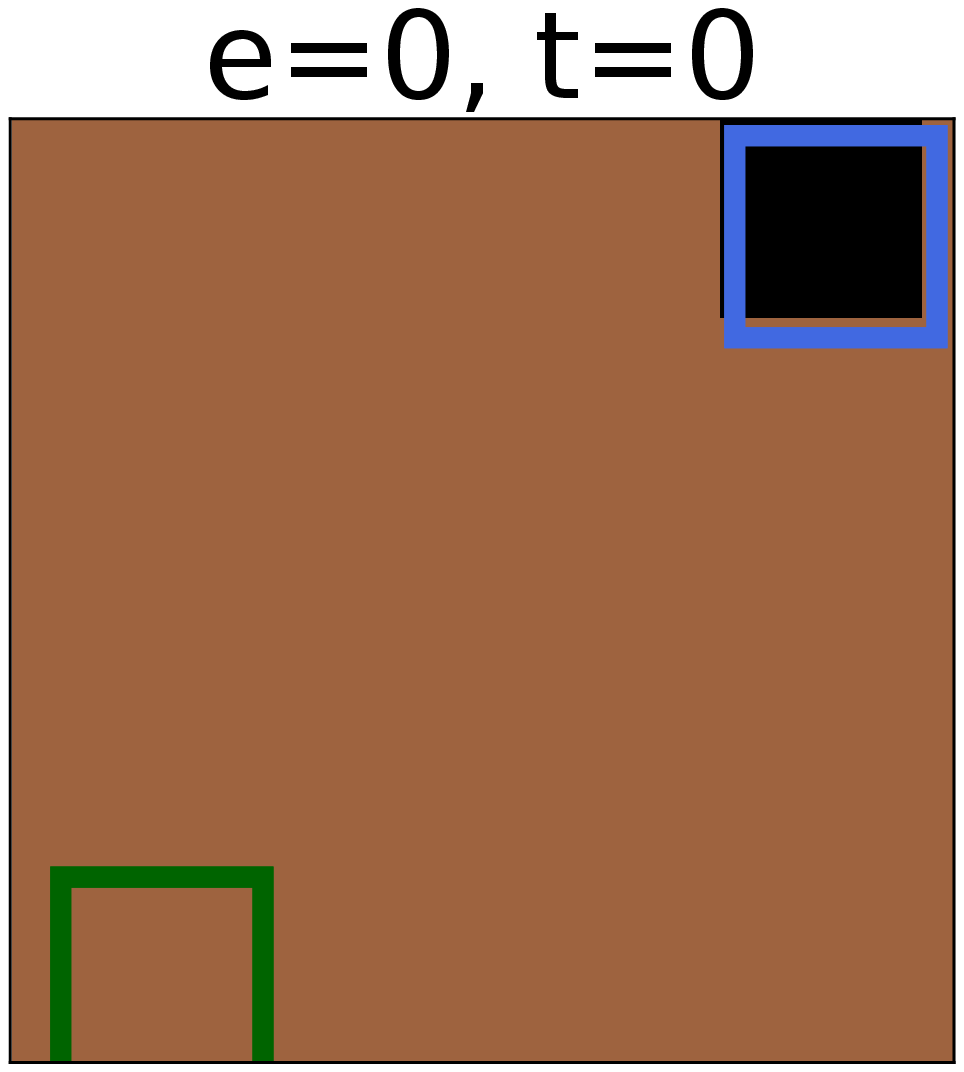}
		\includegraphics[width=2.1cm]{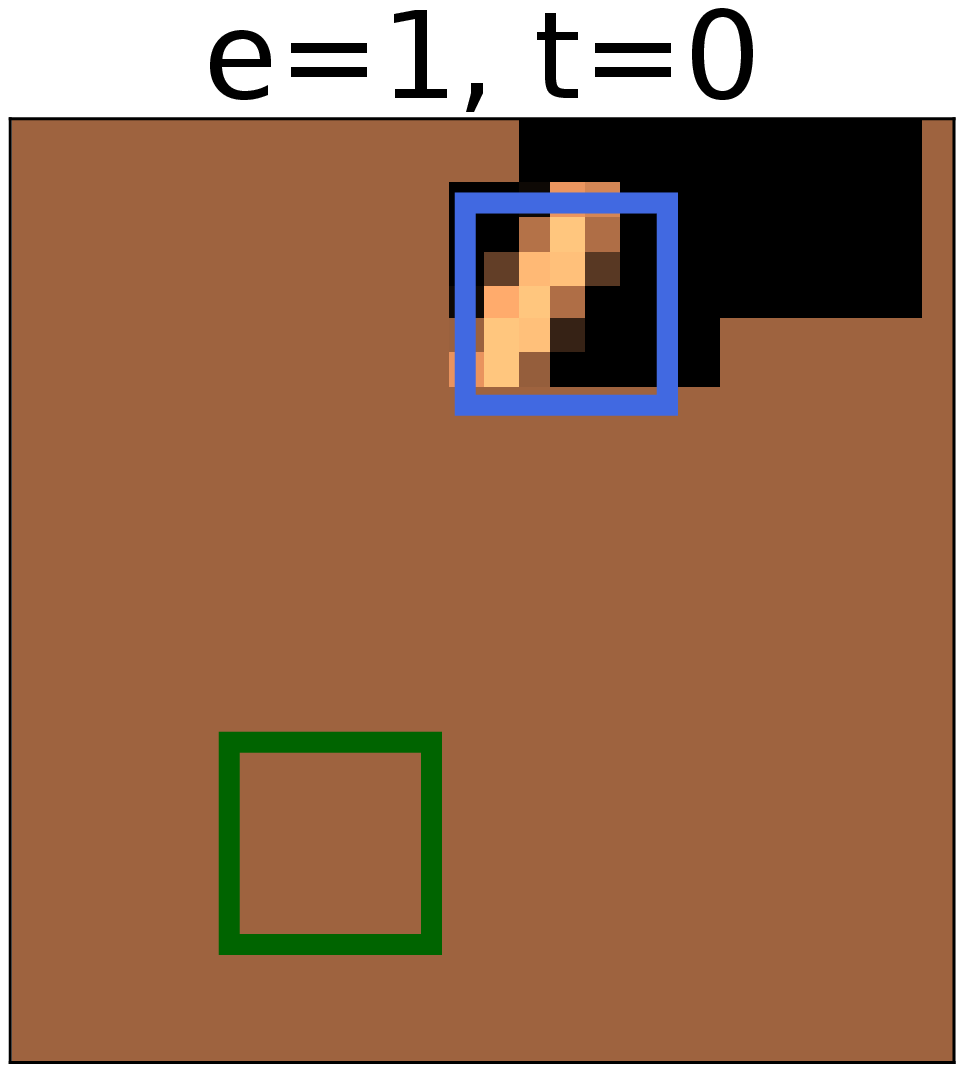}
		\includegraphics[width=2.1cm]{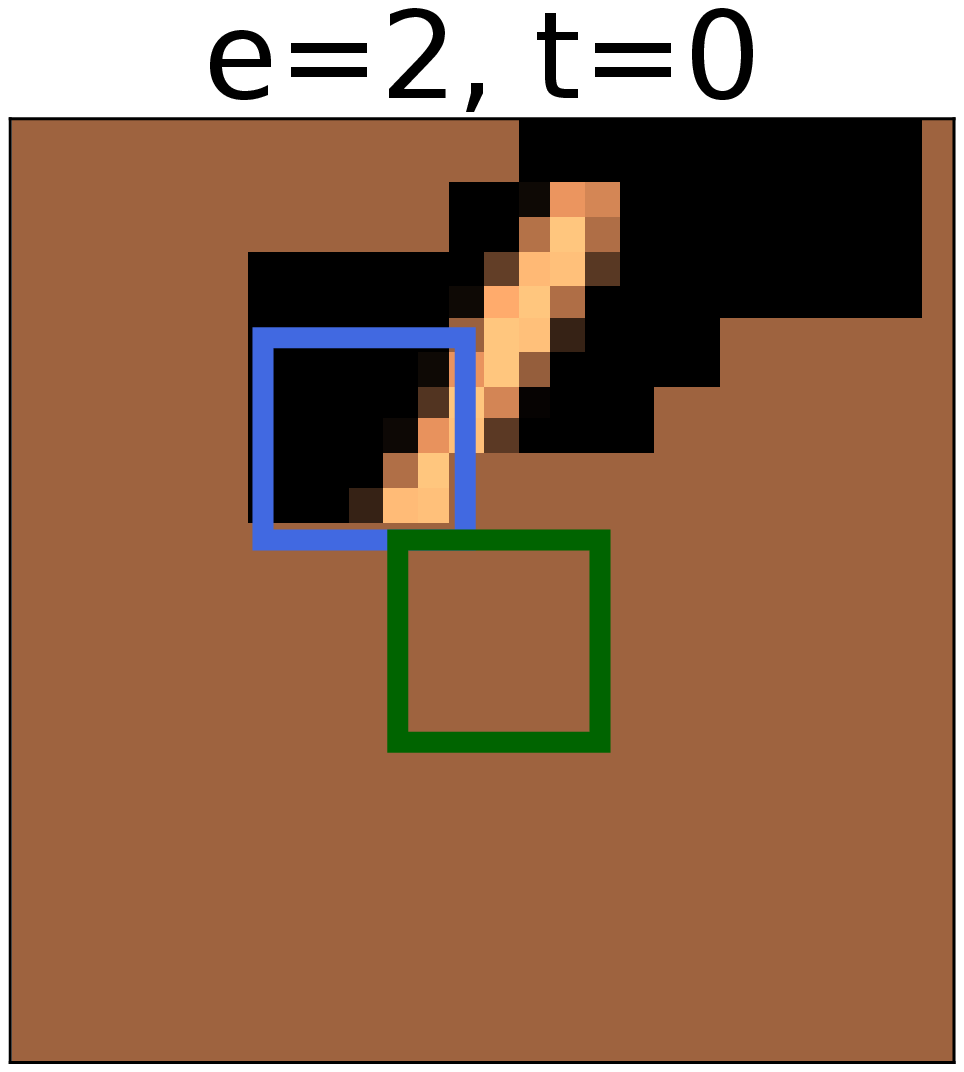}
		\includegraphics[width=2.1cm]{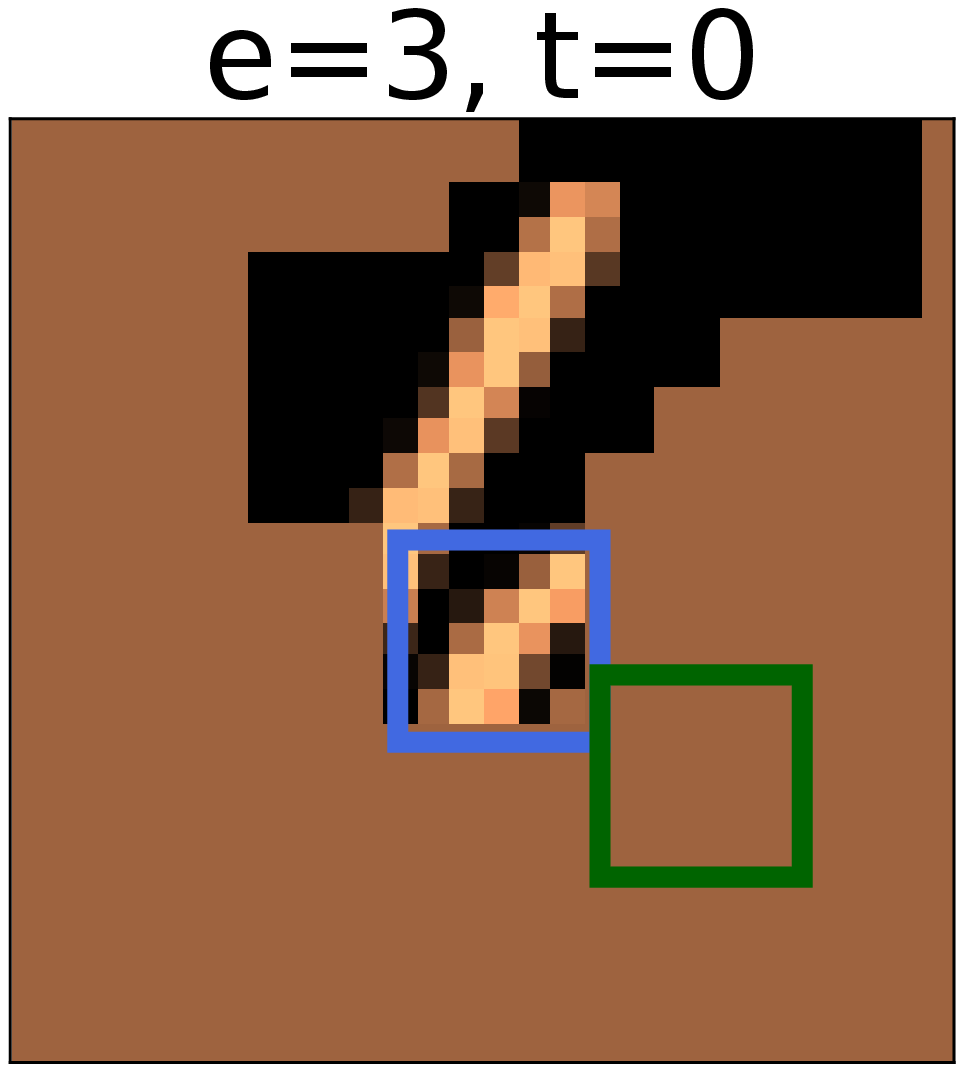}
		\includegraphics[width=2.1cm]{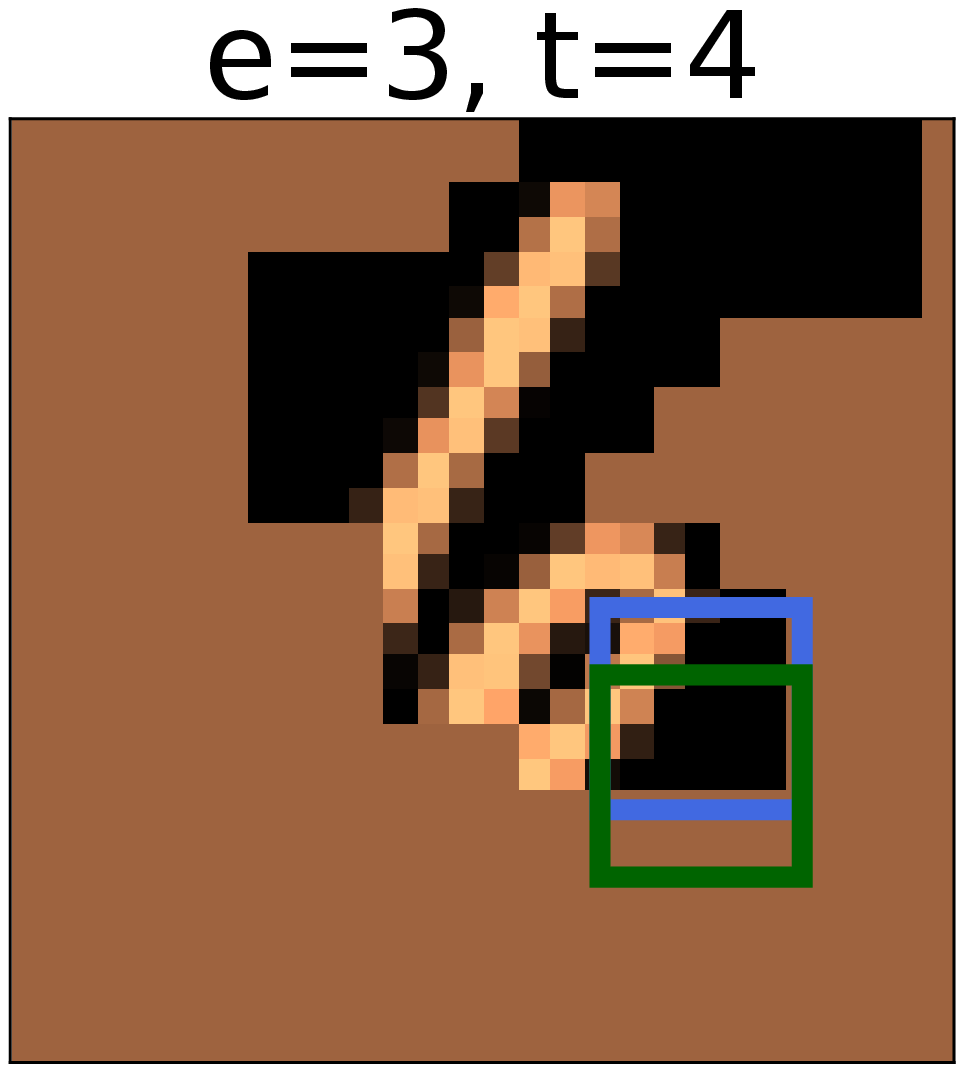}
		\includegraphics[width=2.1cm]{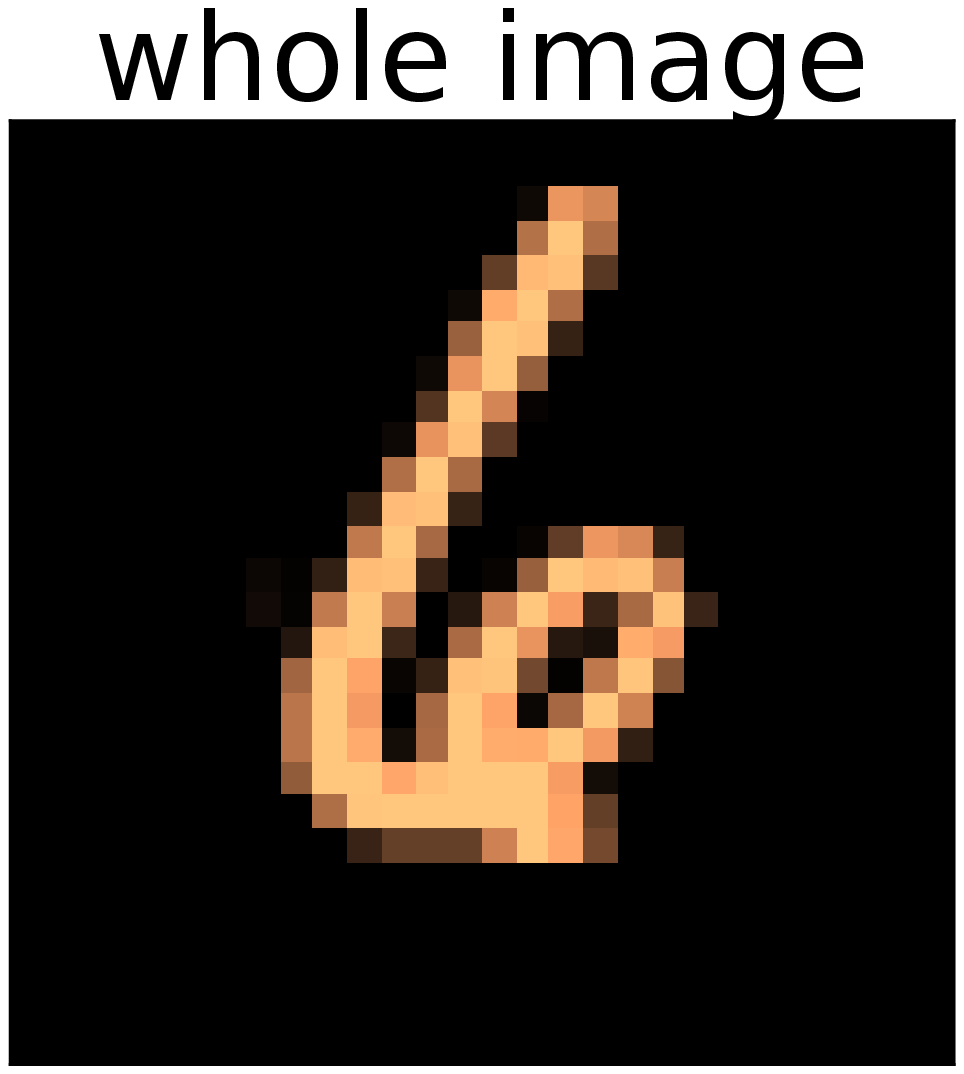}
		\includegraphics[width=2.5cm]{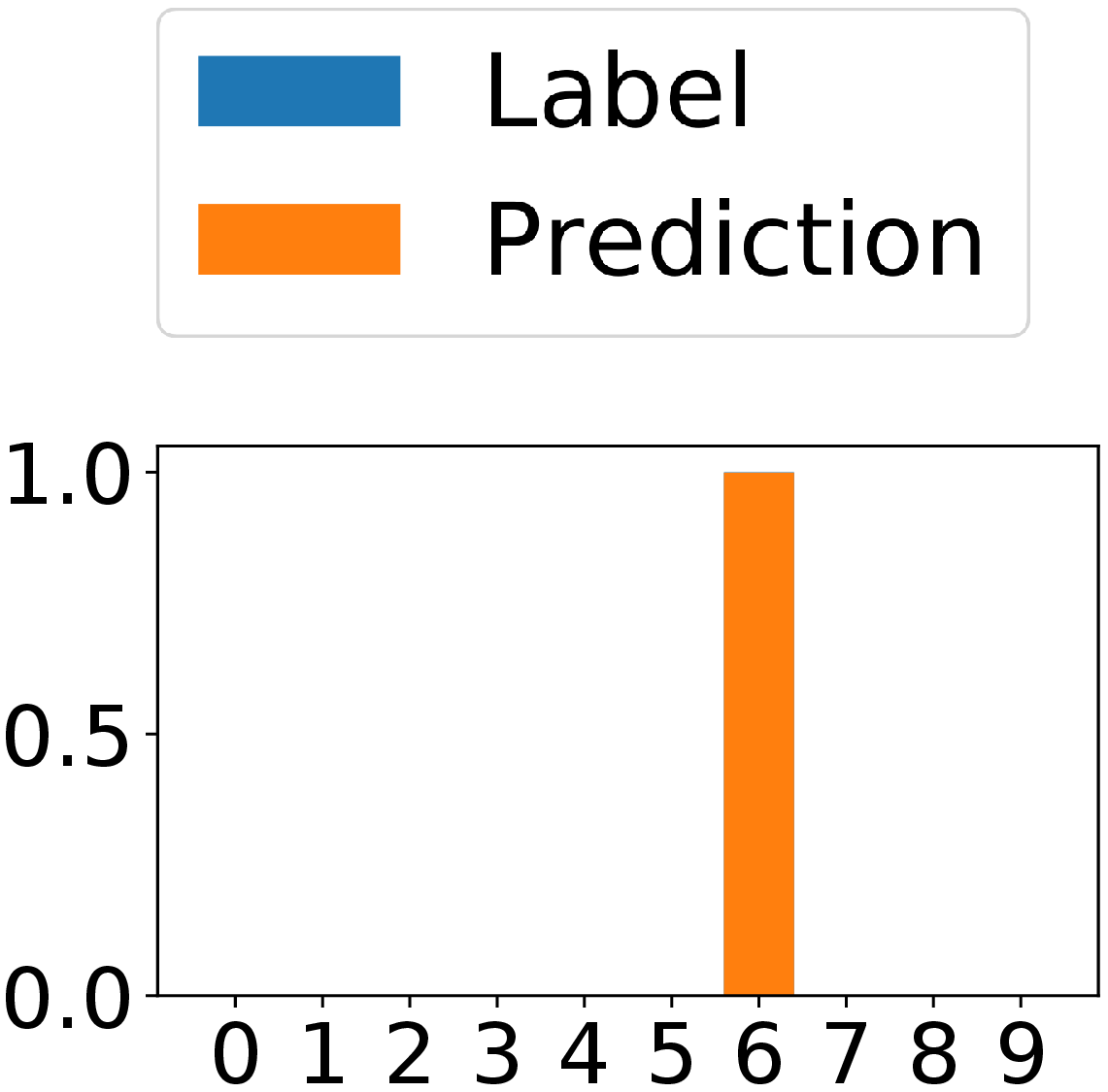}
		
		\includegraphics[width=2.1cm]{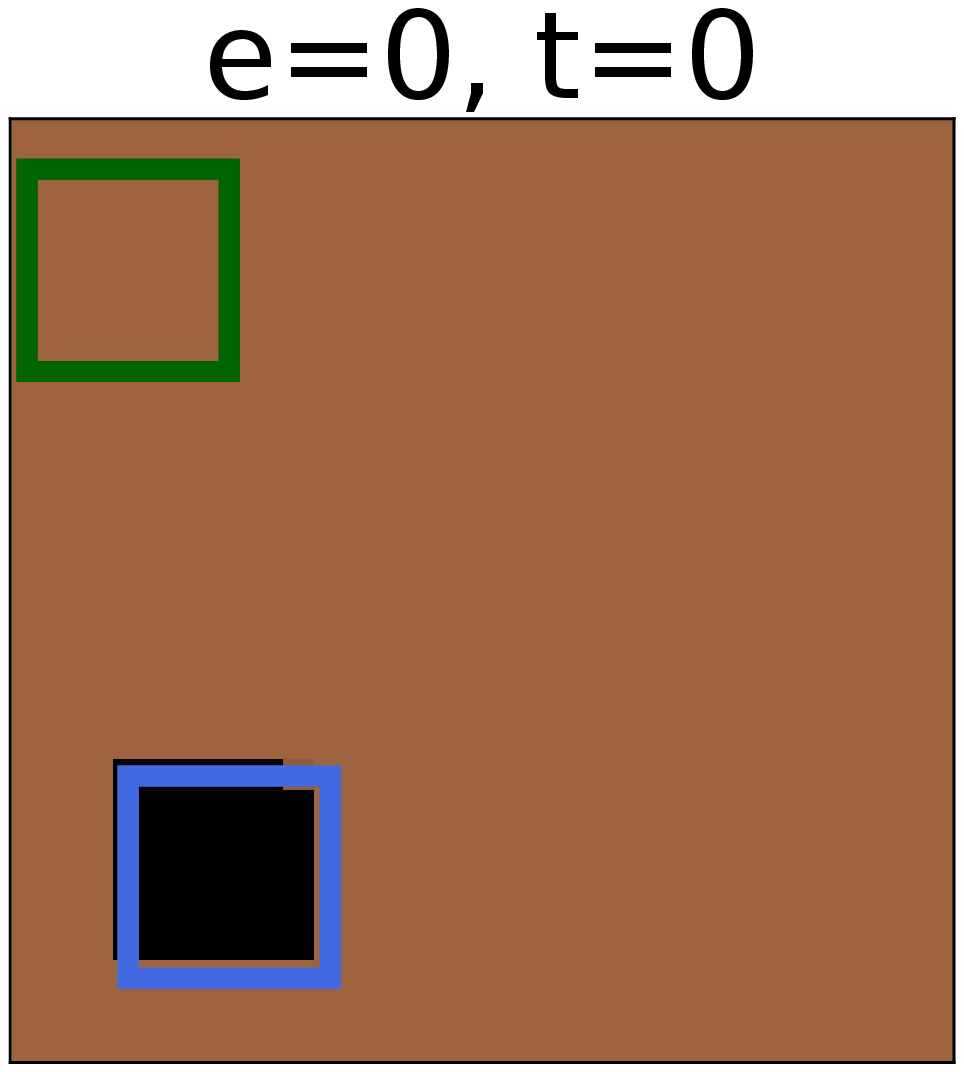}
		\includegraphics[width=2.1cm]{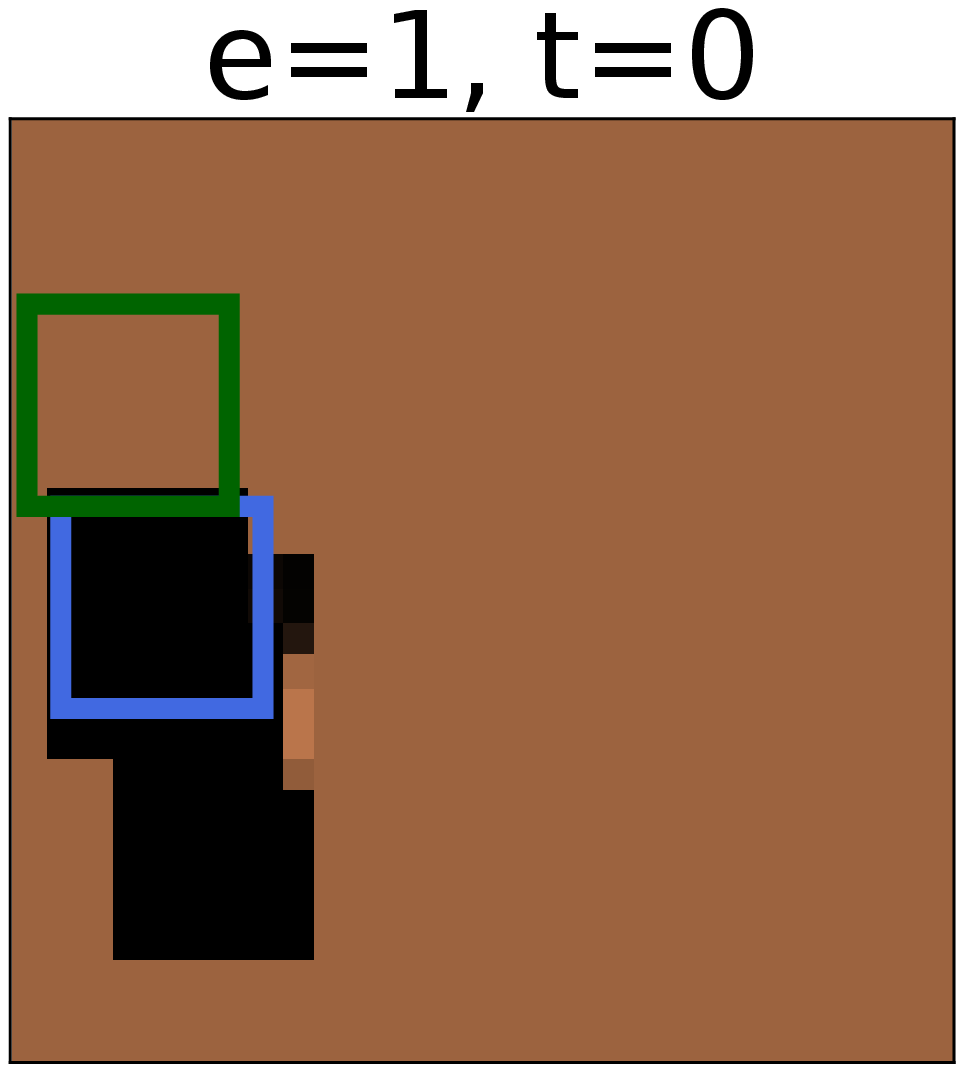}
		\includegraphics[width=2.1cm]{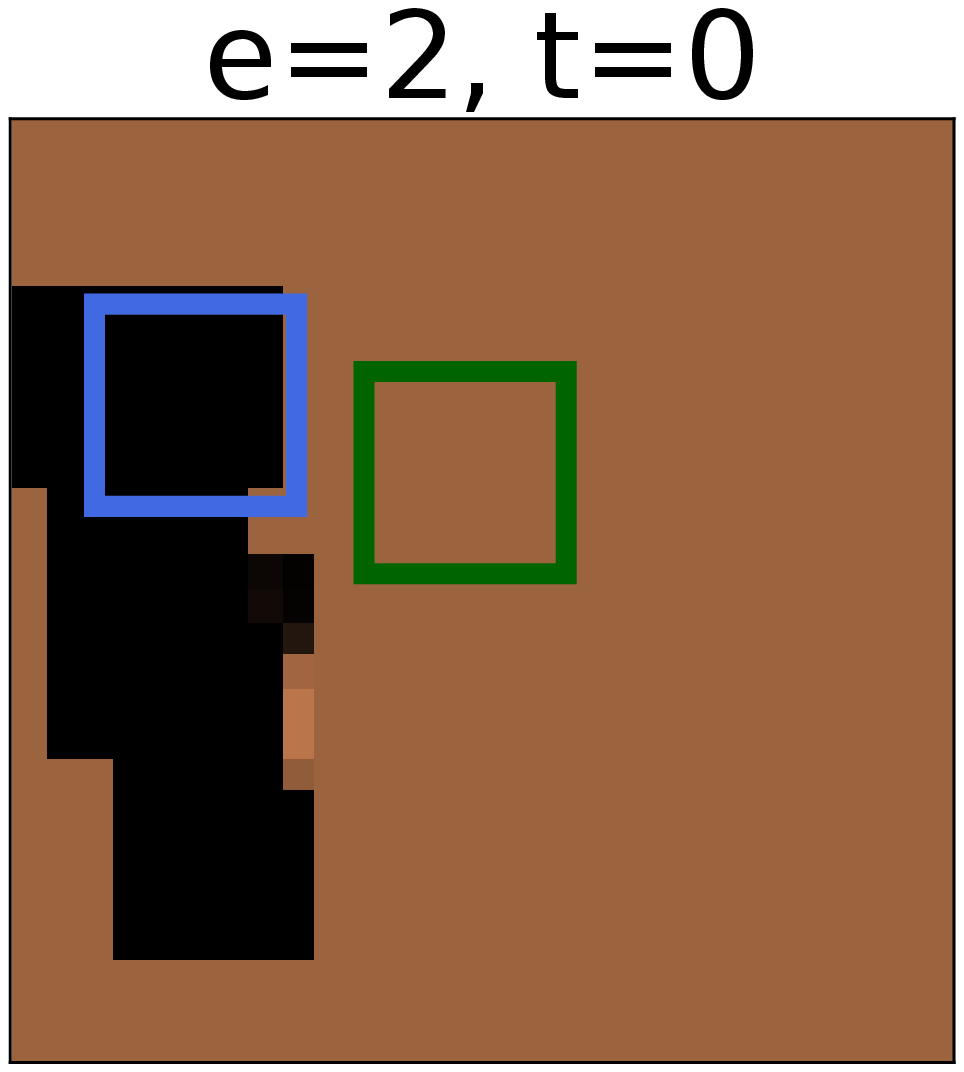}
		\includegraphics[width=2.1cm]{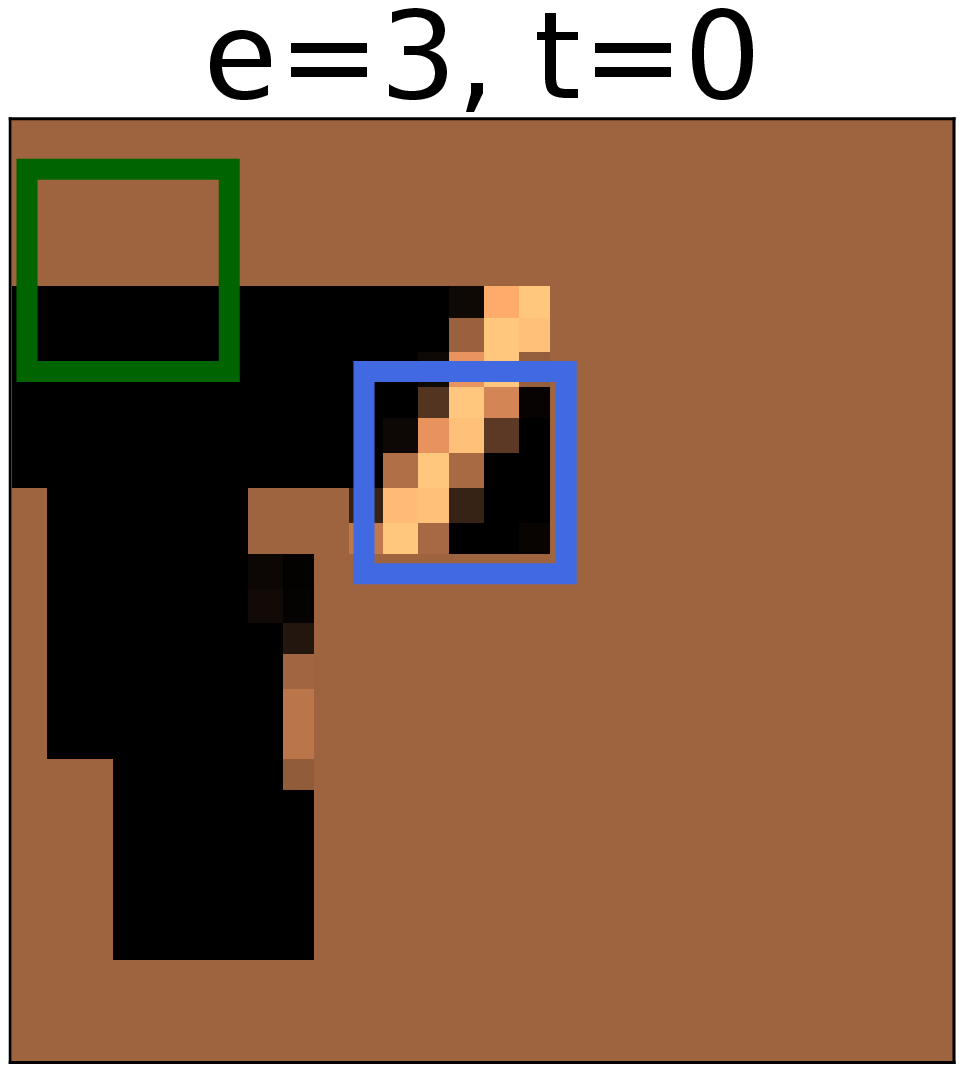}
		\includegraphics[width=2.1cm]{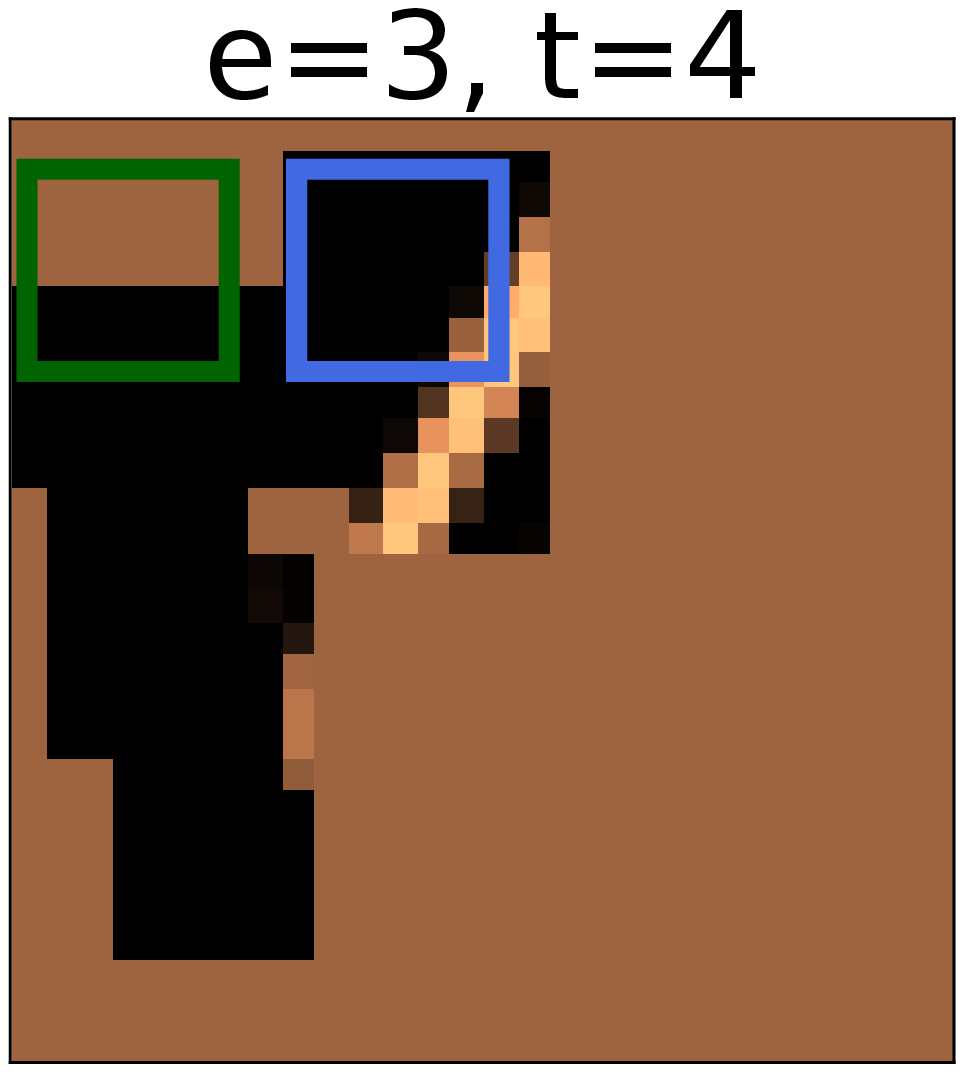}
		\includegraphics[width=2.1cm]{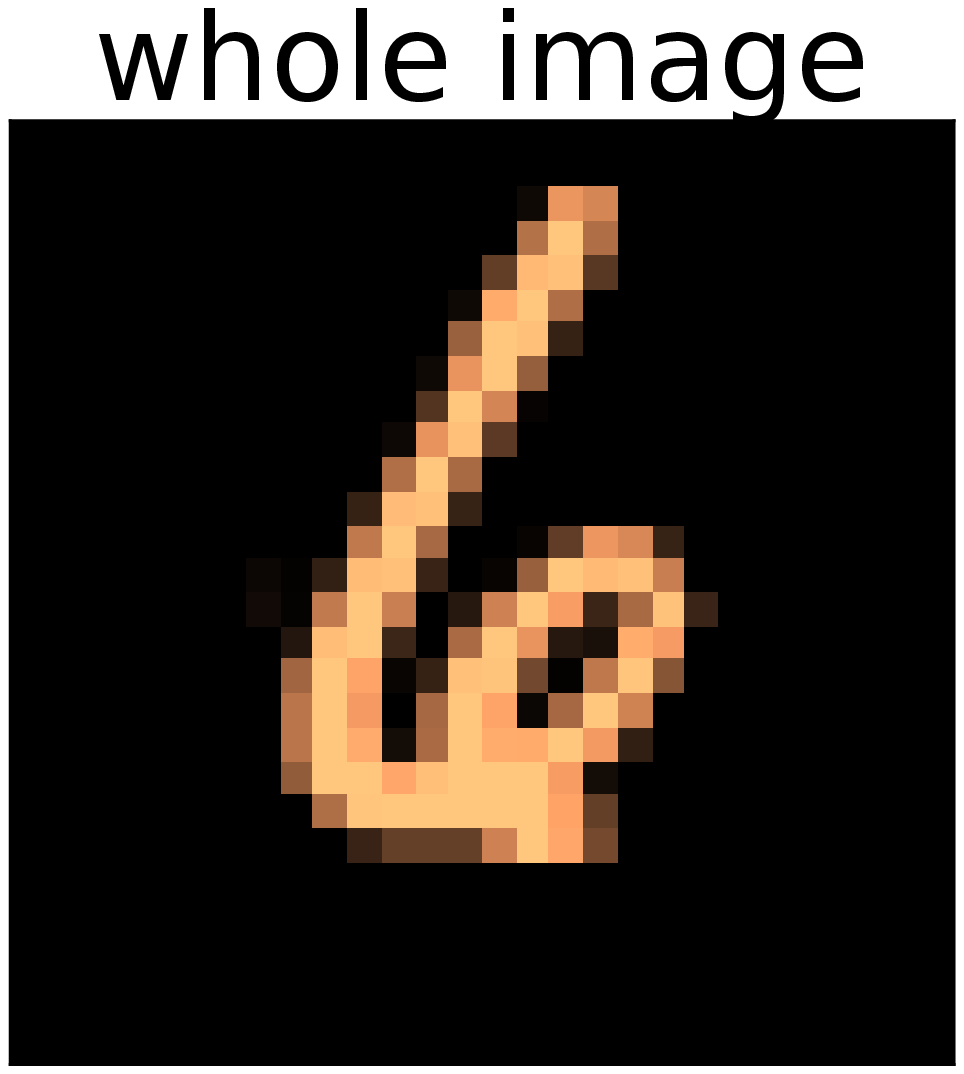}
		\includegraphics[width=2.5cm]{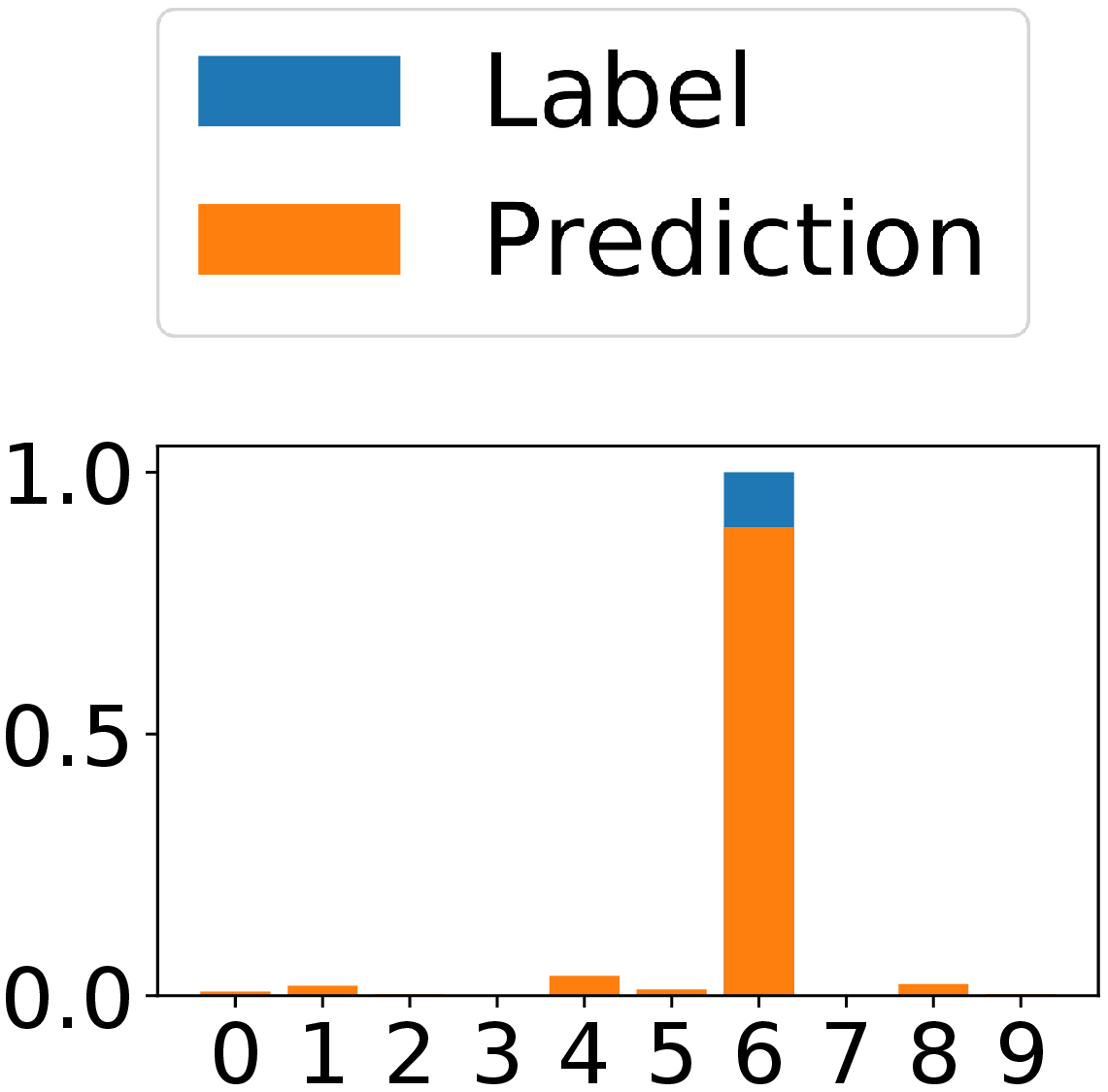}
		
		\includegraphics[width=2.1cm]{images/images/g3/snp0.eps}
		\includegraphics[width=2.1cm]{images/images/g3/snp5.eps}
		\includegraphics[width=2.1cm]{images/images/g3/snp10.eps}
		\includegraphics[width=2.1cm]{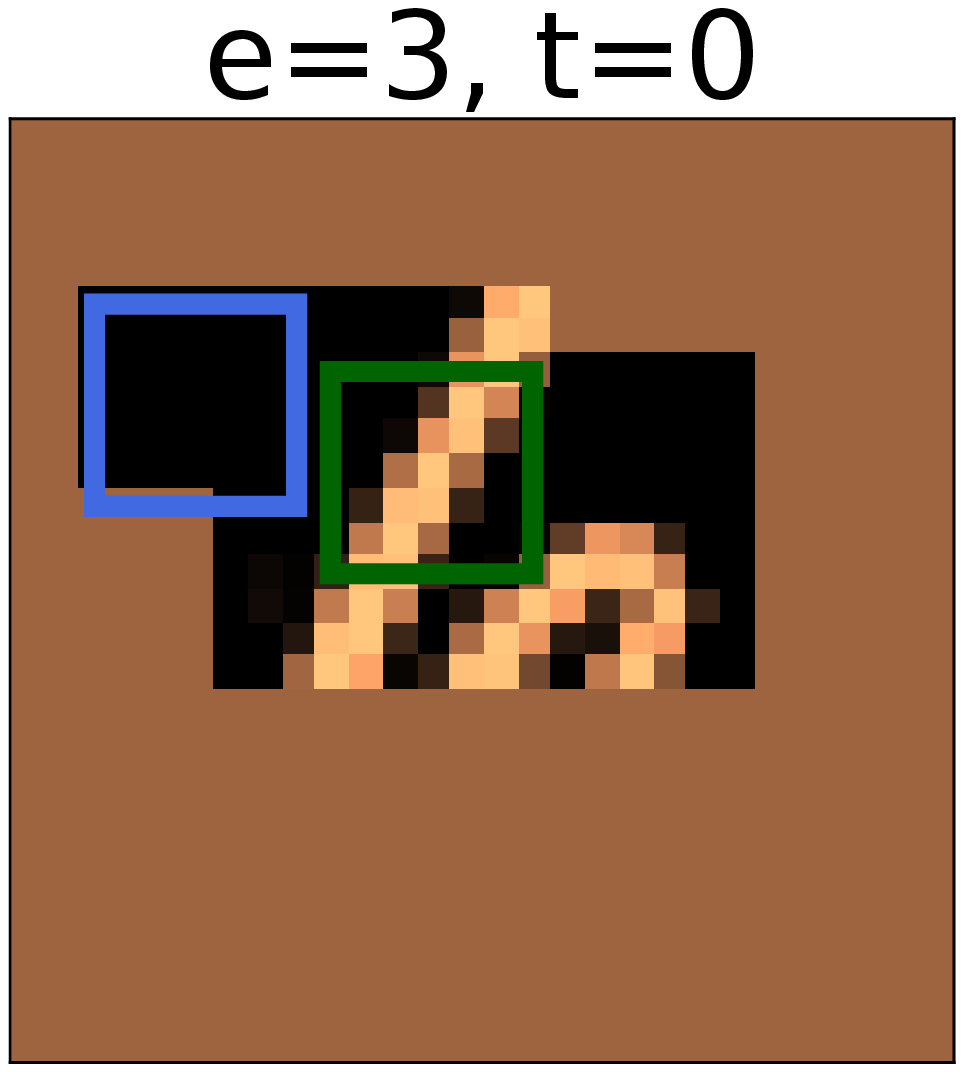}
		\includegraphics[width=2.1cm]{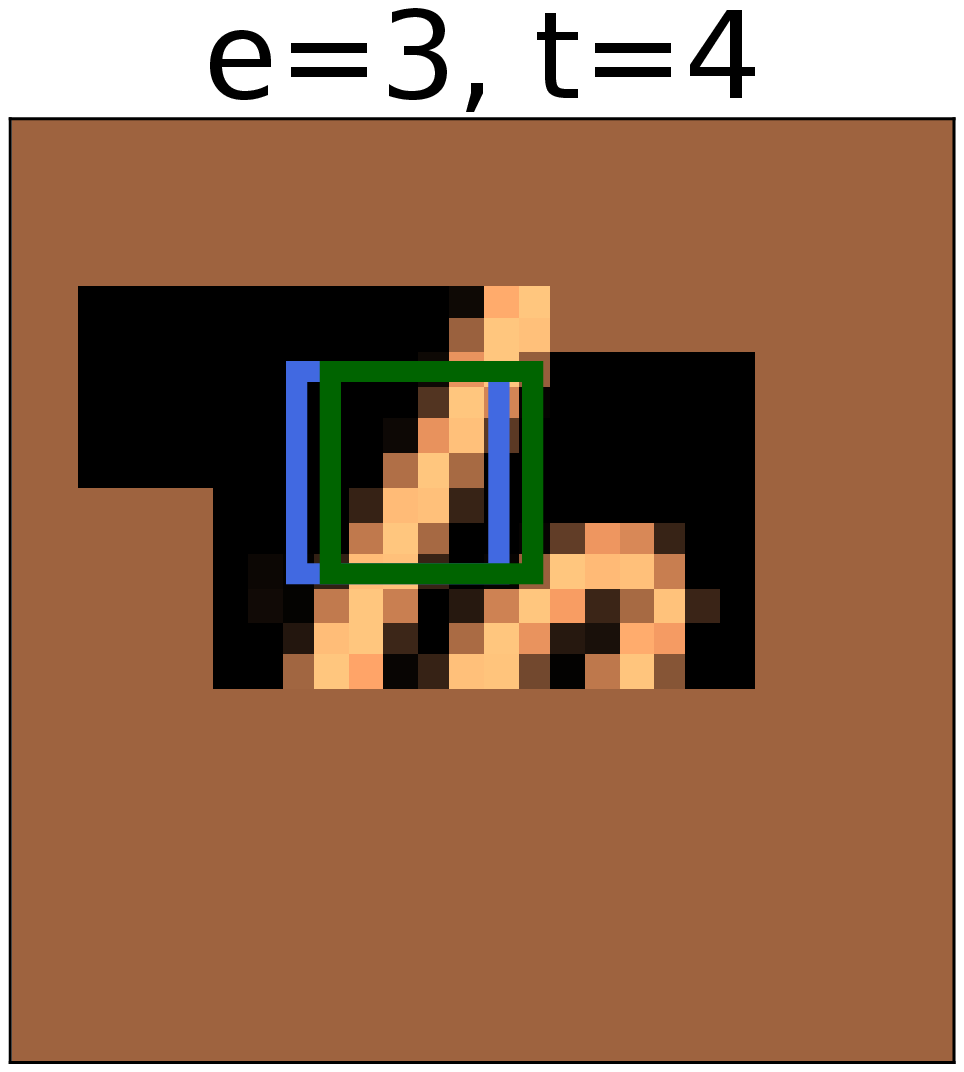}
		\includegraphics[width=2.1cm]{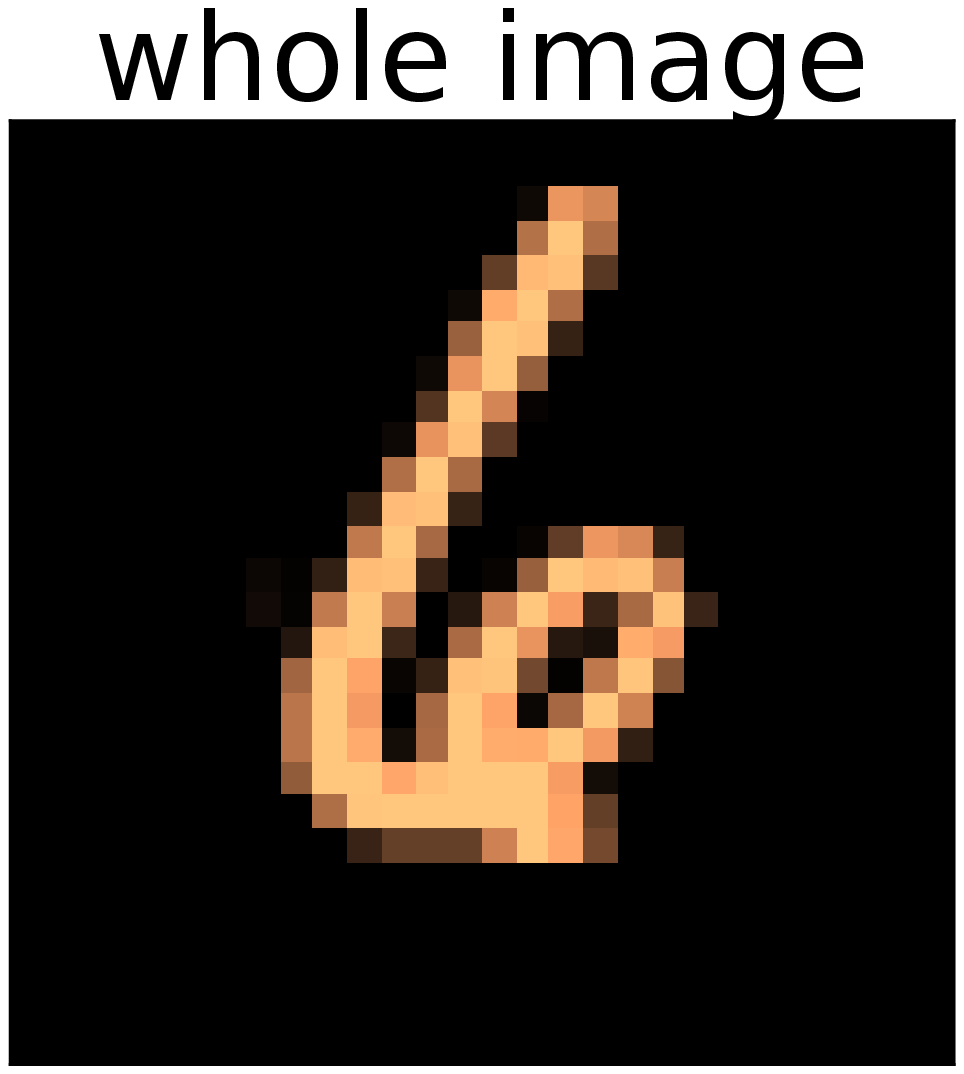}
		\includegraphics[width=2.5cm]{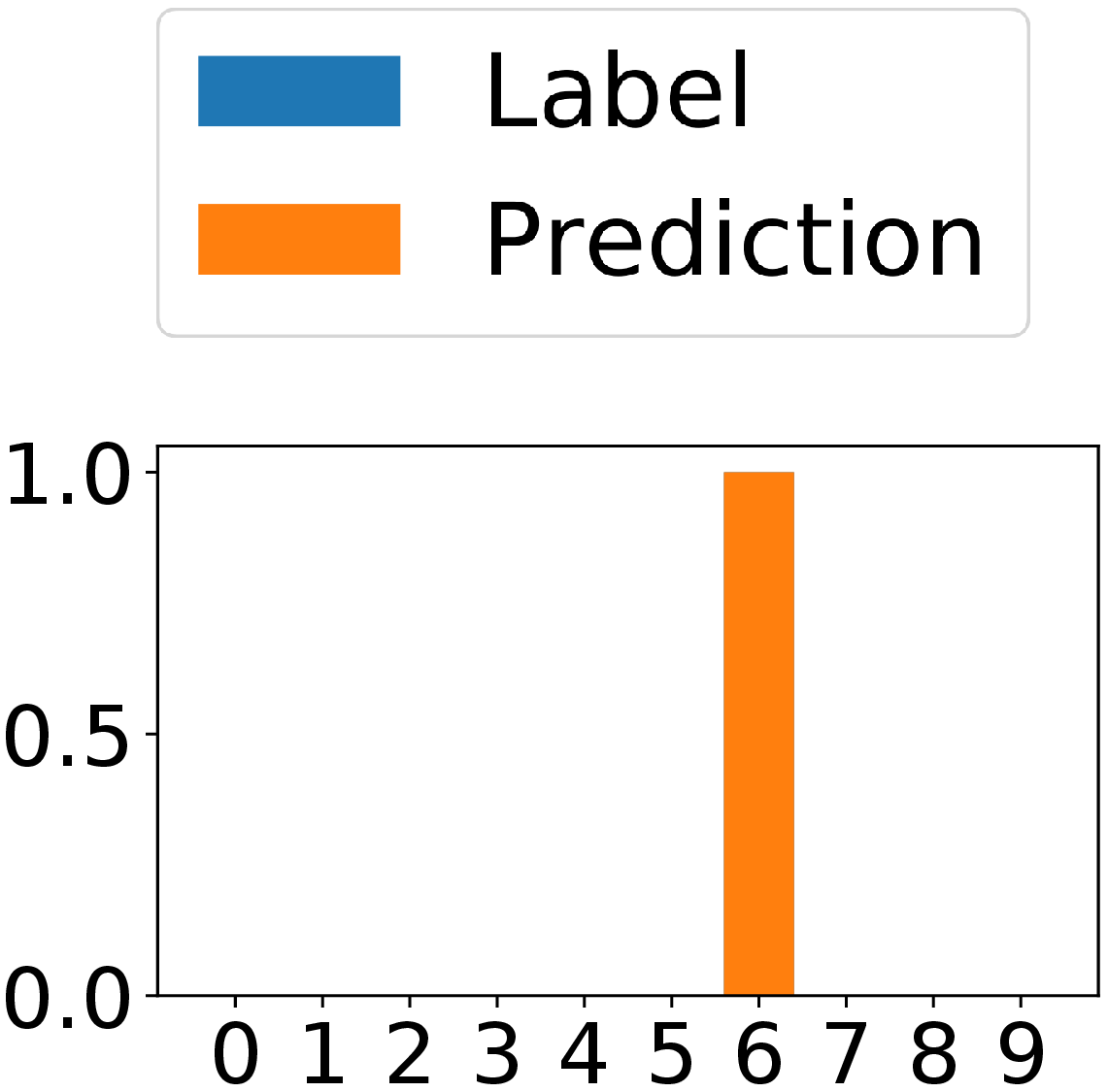}
		
	\end{center}
	\caption{We demonstrate six sample trajectories from two data points. The snapshots have been taken at the beginning of $4$ episodes as well as the end of {the last} episode. The blue and green squares point to the current position and goal position. The prediction corresponding to the final unmasked image is also illustrated in each case. }
	\label{fig:samples}
\end{figure*}

The proposed multi-layered architecture as well as this policy gradient algorithm allow us conduct  training of the three layers (i.e, goal planner, action planner, and classifier) with a wide range of flexibility.  All three modules can be either in training mode or kept fixed after training. Moreover, for goal and action planning layers, we have an extra level of flexibility \emph{before} training: we can consider i.i.d. (i.e., independent and identically distributed) planning of goals or actions. This mode of operation for {goal planner}  implies that the goals are chosen from a uniform distribution over all pixels. This model of operation for action planner means that taking horizontal or vertical actions towards the goal have always equal probability of $1/2$. Once we switch to learning the parameters for either of these planners, we cannot switch back to i.i.d. mode. In Table \ref{table:modes}, we have summarized these possibilities.  In this paper, we consider a sequence of three different training modes:\\
\noindent (i) meta-layer and action-layer in i.i.d. mode, while classifier is being trained,  
\noindent (ii) action-layer in i.i.d. mode, while classifier and goal planner are being trained simultaneously,  
\noindent (iii) all layers being trained simultaneously.

In every mode,  reward $r^\tau$ is equal to $r$  given by \eqref{eq:r}. In mode (i), all goals and actions are identically distributed. Thus, we can arbitrarily set $\log p^\tau=0$ (or any other constant). In mode (ii), only the goals are actively decided. Therefore, the probability term is given by 
$$
\log \pi^\tau=\sum_{e \in [E]}  \log \pi_g(e),
$$ while for mode (iii), we need to set 
$$
\log \pi^\tau=\sum_{e \in [E]} \log \pi_g(e)+\sum_{e \in [E]}   \sum_{t \in [T]}\mathrm{\chi}(e,t)\log \pi_a(e,t),
$$ 
where $\chi({e,t})=1$ if the action at instant $(e,t)$ was decided by  action distribution, and $\chi({e,t})=0$ otherwise.
  


 \begin{figure}[t]
	\begin{center}
		\includegraphics[width=7cm]{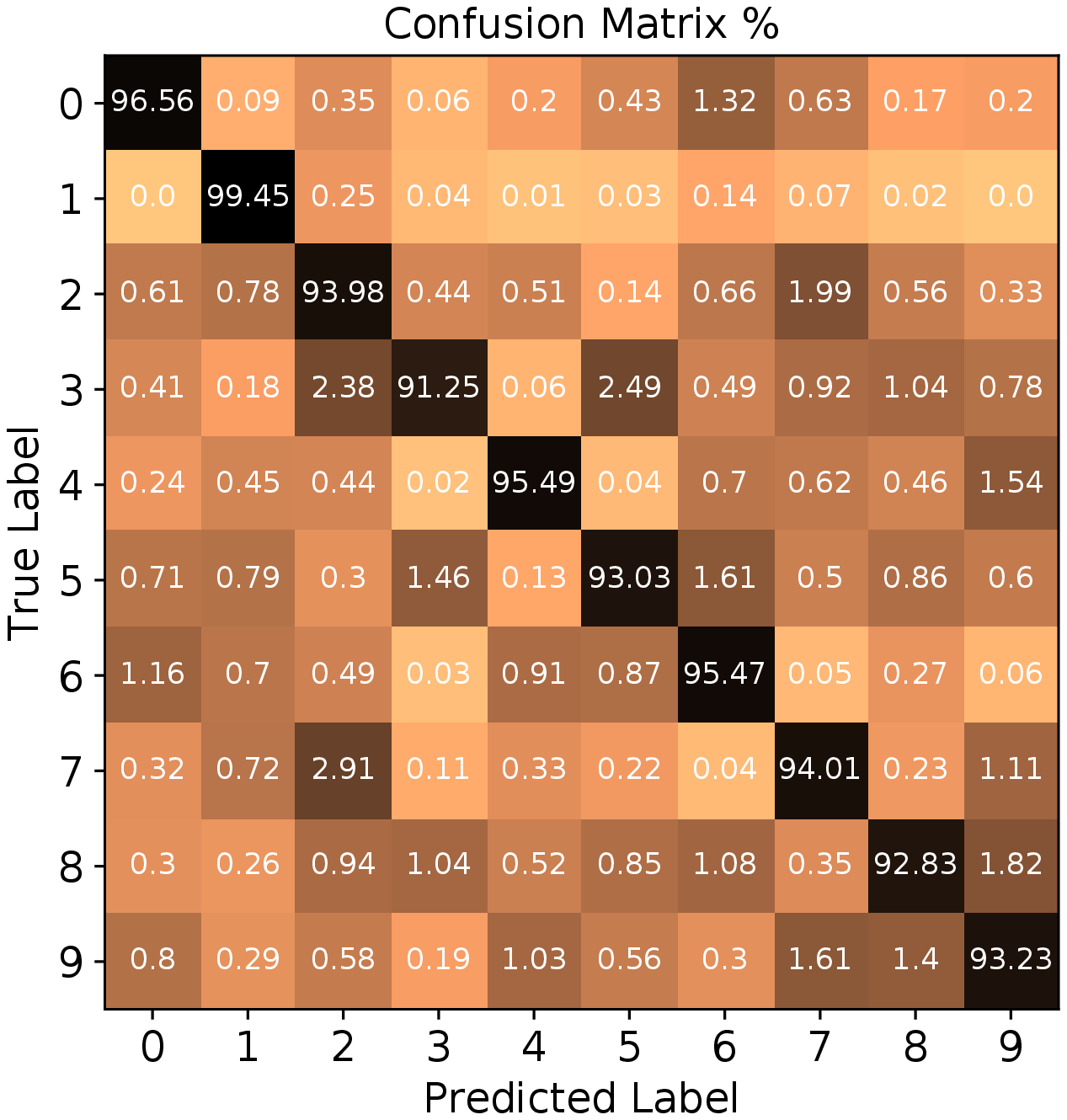}
	\end{center}
	\caption{The confusion matrix of  classification computed on the test dataset. The reported numbers are averaged over $20$ runs of data.  }
	\label{fig:confusion}
\end{figure}

 \begin{figure}
	\begin{center}
		\includegraphics[width=6.8cm]{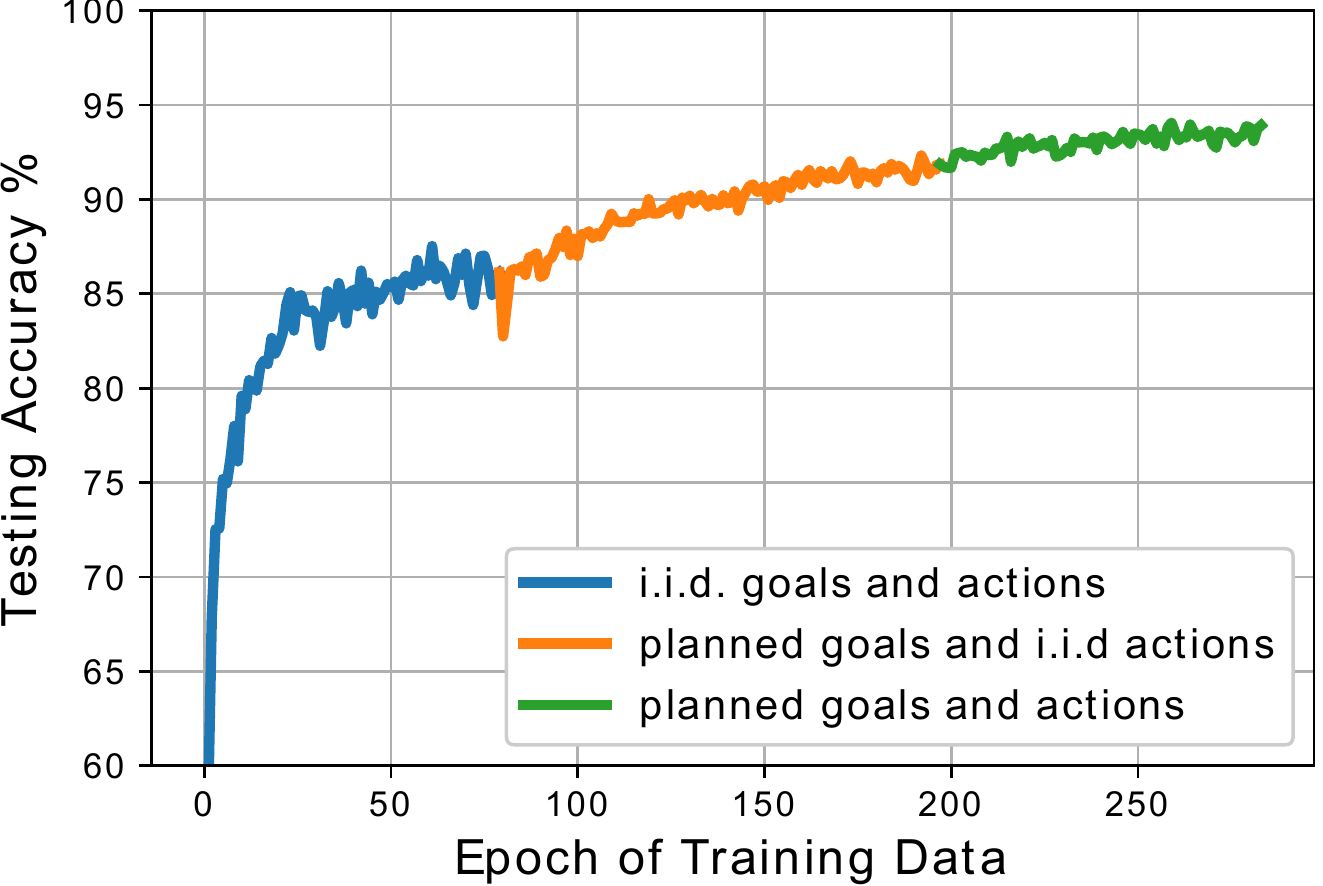}
	\end{center}
	\caption{The testing data accuracy vs. data epoch using the hierarchical training sequence from scratch. }
	\label{fig:progress}
\end{figure} 

\begin{figure}
	\begin{center}
		\includegraphics[width=7.0cm]{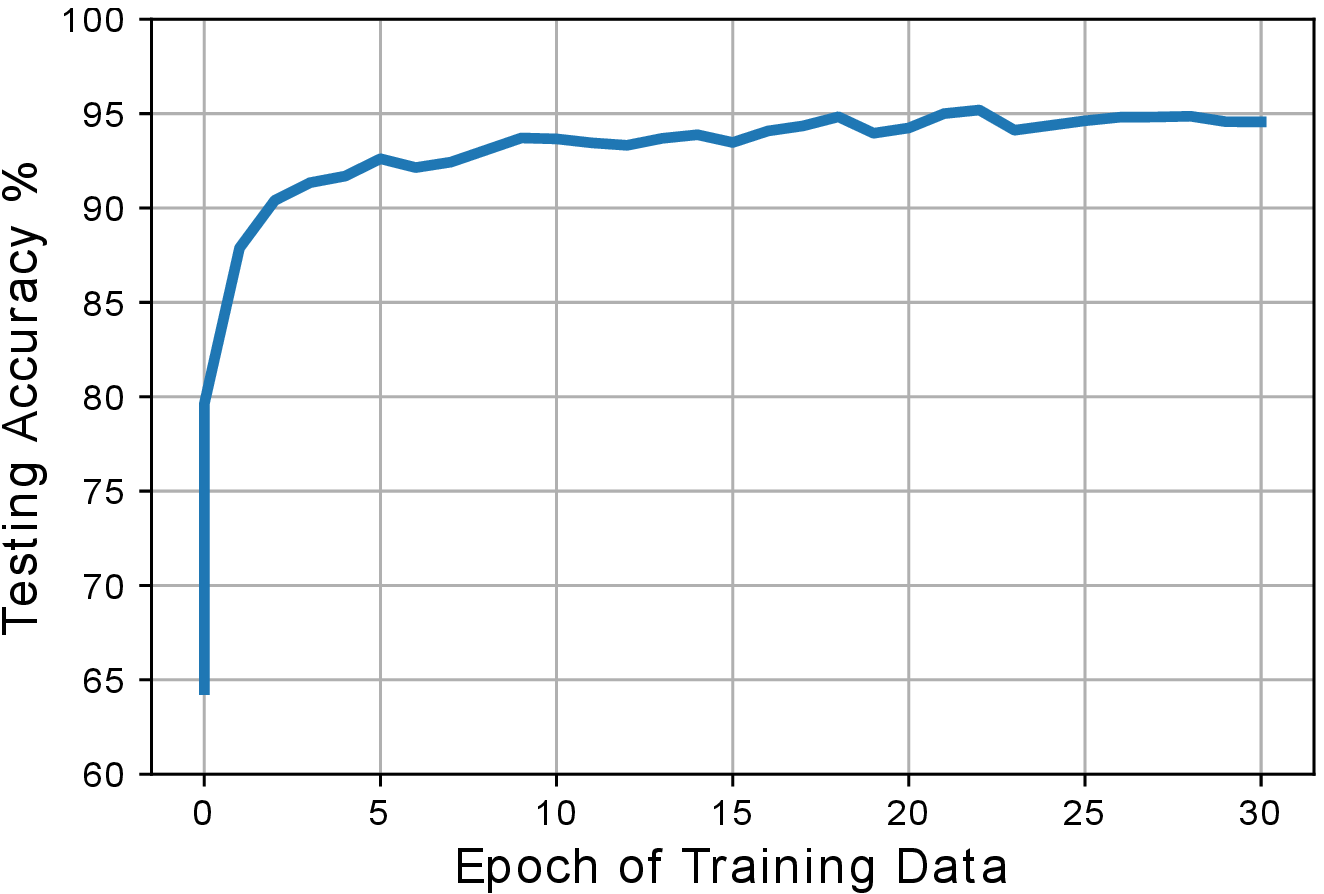}	
	\end{center}
	\caption{Testing data accuracy vs. data epoch with transfer learning. }
	\label{fig:progresstf}
\end{figure}

 \section{Numerical Experiment}

We test the method on {the} MNIST dataset of handwritten digits \cite{lecun1998gradient}. The dataset consists of $60,000$ training examples and $10,000$ test images, each of $28 \times 28$ pixels.

\noindent{\bf General Setup:} The dataset was normalized between $-0.5$ and $0.5$.   
In all experiments, the agent starts at a random {position} inside the image. The actions in any direction are $2$ pixels per step. In these experiments, we did not use the test set for hyper-parameter tuning. We used student's $t$-test for the confidence interval of  stochastic accuracies  with $\alpha$-value of $5\%$. The number of rollout per data point was 4 in the experiments (unless  otherwise). We used Adam solver for the optimization with a mini-batch size of $60$ images.  The model was built in PyTorch \cite{paszke2017automatic}.

\noindent{\bf Sample Accuracy Results: } We conduct the training with patch size $m=6$ for $E=4$ episodes that each have a horizon of $T=5$. The training and testing accuracies for the trained model where $94.39 \pm 0.03\%$ and $94.61\pm 0.17\%$, respectively. This suggests an acceptable level of generalization for our trained model to unseen test set, while the accuracy on the test set has a slightly higher variance.  

 \noindent{\bf Sample Trajectories: } In Fig. \ref{fig:samples}, we demonstrate $3$ sample trajectories on $2$ test data points next to resulting prediction probabilities.  We have intentionally illustrated both high confidence and low confidence outcomes. For instance, on the test point with label $4$, the second trajectory results in a wrong prediction, which is likely due to the fact that the agent has not uncovered the upper region of $4$ in its limited temporal budget. As one observes, in most cases, the goals and actions are selected such that the agent can see the most informative parts of the image. 

\noindent{\bf  Top Two Category  Accuracy: } For the previously described model, we evaluate the top $2$ class accuracy (i.e., if the true label is among the top $2$ categories predicted by the model). Then, the training and testing accuracies increased to $98.27 \pm 0.02\%$ and $98.30\pm 0.04 \%$, respectively.

\noindent{\bf Confusion Matrix:} For the trained model, we build the confusion matrix of the classification for the testing data. In Fig. \ref{fig:confusion}, we show this matrix. The reported accuracies are averaged over  $20$ independent experiments.

\noindent{\bf Performance of Classifier Module:} Let us consider  the trained classifier module with complete (i.e, unmasked) image as its input; i.e., $u_c=x$. We can evaluate the performance of this isolated model, which turns out that the training and testing accuracies were $94.85\%$ and $94.92\%$, respectively. The accuracies for top two categories for the classifier module were $98.32\%$ and   $98.52 \%$ on the training and test sets, respectively. This suggests that the planning layers (meta-layer and action-layer) are successfully revealing the most informative regions of the image.

\noindent{\bf Accuracy Vs. Epoch:} In Fig. \ref{fig:progress}, we demonstrate the testing accuracy versus training epochs, which is based on
  hierarchical training sequence that was described in Subsection \ref{subsec:hi}. The  introduces two random baselines in addition to the final model: the model in which the goals and actions are decided in i.i.d. manner, in addition to the model in which the goals are planned, but the actions are planned in i.i.d. manner. Fig. \ref{fig:progress} reveals that the errors in prediction have decreased by around $1/3$ after using the goal planner, and by almost another $1/3$ after incorporating the action planner. 
  
 \noindent{\bf Transfer Learning:} In the previous experiment, all classification and planning layers  were trained   from scratch. However, transfer learning ideas suggest that we may accelerate training if we can pretrain some modules. To this end, first, we consider ResNet-18 architecture and pretrain it on the the dataset (with full images) for $15$ epochs. This resulted in more than $99\%$ testing accuracy on the full images. Then, we replace the classification architecture in our system with ResNet-18 and start training \emph{all layers} (i.e., planning and perception). The result of training is illustrated in Fig. \ref{fig:progresstf}, which shows that the maximum testing accuracy of $95.19\%$ was acheived in a considerably shorter period of training (by almost an order of magnitude). 


  \section{Concluding Remarks} 
  
We introduced a three-layer architecture for active perception of an image that allow us to co-design planning layers for goal generation and local navigation as well as  classification layer. The layered structure of the proposed  mechanism and the unified definition of reward for all layers enable us to train the parameters of the deep neural networks using a policy gradient algorithm. We would like to discuss a number of final remarks. 
  
First,  we did not use any overfitting prevention measures (dropouts, weight decay, etc.) in our models. However, even without use of validation sets, we observe a very good level of generalization of the current model. This may be explainable by use of fully-convolutional layers and global average pooling before evaluating the probability vectors, as suggested by  \cite{lin2013network}.
  
  Second,  
   %
     variations of the current architecture with recurrent memory (e.g., LSTM cells as used in \cite{mousavi2019multi}) are straight-forward to construct. {This could be particularly useful when we extend our results to multi-robot scenarios.} 
      
Third,  the intrinsic partial observability of this problem motivates use of policy gradient algorithms rather than Q-learning approaches \cite{kulkarni2016hierarchical}. It is an interesting line of research to develop Q-learning techniques that perform at the same level as the sampling based approaches for this class of problems.

%

	

	\bibliographystyle{IEEEtran}
	
	\bibliography{mybib}

\end{document}